\def\ICLRPublicVersion{1}
\def\ICLRArxivVersion{1}
\documentclass{article} 
\usepackage{iclr2027_conference,times}

\usepackage{graphicx}
\usepackage{amssymb}
\usepackage{mathtools}
\usepackage{booktabs}
\usepackage{array}
\usepackage{float}
\usepackage{placeins}
\usepackage{needspace}
\usepackage{algorithm}
\usepackage{algorithmic}
\usepackage{url}
\usepackage[hypertexnames=false]{hyperref}
\hypersetup{hidelinks}

\title{Reliability-Aware Checkpoint Selection for Domain Generalization}

\ifdefined\ICLRPublicVersion
\author{%
  \textbf{Jinshi Liu}$^{1,*}$ \quad
  \textbf{Jiahao Li}$^{2,*}$ \quad
  \textbf{Pan Liu}$^{3}$ \quad
  \textbf{Yanfeng Li}$^{4}$ \\
  \textbf{Rui Qian}$^{5,\dagger}$ \quad
  \textbf{Zhao Tong}$^{6}$ \quad
  \textbf{Yue Sun}$^{4}$ \quad
  \textbf{Tao Tan}$^{4}$ \\[0.35em]
  \normalfont\small
  $^{1}$Shenzhen University \quad
  $^{2}$Xiamen University \\
  $^{3}$The Hong Kong University of Science and Technology (Guangzhou) \\
  $^{4}$Macao Polytechnic University \quad
  $^{5}$Fudan University \\
  $^{6}$Institute of Information Engineering, Chinese Academy of Sciences \\
  $^{*}$Equal contribution. \quad $^{\dagger}$Corresponding author.%
}

  \iclrfinalcopy
\else
  \author{Anonymous Authors}
\fi

\begin{document}

\maketitle

\ifdefined\ICLRArxivVersion
  \lhead{Preprint.}
\fi

\begin{abstract}
Checkpoint selection in domain generalization often relies on source-validation accuracy, yet the selected checkpoint need not provide reliable probabilities on unseen target domains.
Source--target distribution shifts can alter accuracy rankings, while accuracy alone does not measure predictive probability quality.
We identify an empirical selection opportunity within fixed training trajectories: reselecting among checkpoints with near-optimal source accuracy can improve mean target probability quality with small observed changes in mean target accuracy.
We study accuracy-constrained reliability selection (AC), which retains checkpoints within a tolerance of the best source-validation accuracy and ranks them by source reliability. Our reference rule aggregates within-set normalized negative log-likelihood (NLL) and class-wise calibration error (CwECE) using \(D_\infty\).
AC uses no target data and requires neither additional training nor weight averaging.
We evaluate five domain generalization training algorithms on three benchmarks, using PACS to develop the objectives and a 0.5-percentage-point tolerance.
In exploratory aggregation comparisons on 360 OfficeHome and TerraIncognita runs, the reference rule reduces mean target soft-bin squared-gap ECE and CwECE by 0.240\% and 0.182\%, respectively, and NLL by 0.030 relative to Source-Acc.
Mean target accuracy changes by +0.213 percentage points.
These results identify opportunities for reliability-aware reselection, while the additional benefit of joint over single-objective ranking remains unresolved.
\ifdefined\ICLRPublicVersion
Code is available at \href{https://github.com/Jjjjjjh666/Reliability-Aware-DG}{\textcolor{magenta}{GitHub}}.
\fi

\end{abstract}

\section{Introduction}
Domain generalization (DG) trains on multiple source domains for deployment on an unseen target domain. Even within a fixed training run, deployment requires choosing a checkpoint without target-domain validation data. Under the DomainBed training-domain validation protocol, Source-Acc selects the checkpoint with the highest mean source-validation accuracy \citep{gulrajani2021in}. We ask whether the same held-out source data can also distinguish checkpoints by predictive reliability.

\begin{figure}[t]
\centering
\includegraphics[width=\linewidth]{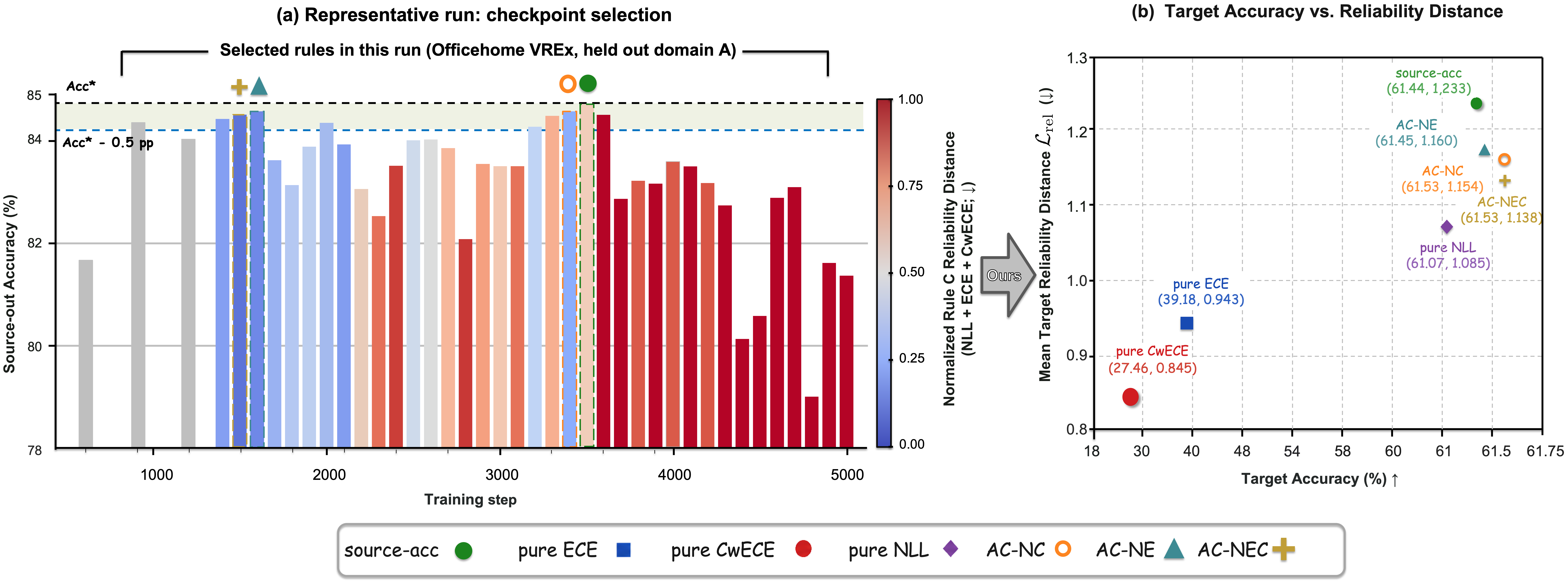}
\caption{Checkpoint selection and the accuracy--reliability trade-off.
(a) A representative OfficeHome/VREx trajectory, with the source-accuracy
tolerance marked and checkpoints colored by source reliability.
(b) Mean target accuracy versus a diagnostic target reliability distance
across the pooled 540 trajectories, including PACS development, with the
high-accuracy region magnified. Objective-set abbreviations are defined in
Section~\ref{sec:ac}. The plotted target distance is not AC's source-side selection
score. Panel (b) is descriptive; target metrics are used only for evaluation.}
\label{fig:selection-overview}
\end{figure}

Accuracy alone ignores probability quality: checkpoints can make identical class predictions while assigning different probabilities. We measure reliability using NLL and calibration errors (ECE and CwECE), motivated by known calibration and uncertainty degradation under distribution shift \citep{guo2017calibration,ovadia2019can}. Reliability-only selection is unsafe: minimizing source ECE or CwECE can select low-accuracy early checkpoints, while unconstrained NLL provides no explicit accuracy budget. Yet near-best-accuracy checkpoints can still differ substantially in reliability (Figure~\ref{fig:selection-overview}).
Following accuracy-filtered calibration selection\citep{wald2021calibration}, we study reliability-based ranking among near-best-accuracy checkpoints within a fixed training trajectory.

We instantiate this principle as accuracy-constrained reliability selection (AC). AC retains checkpoints within \(\delta\) percentage points of the best source-validation accuracy; AC-NC is our reference rule using normalized NLL and CwECE with \(D_\infty\). It returns a saved checkpoint without target data, additional training, or weight averaging. The tolerance controls empirical source accuracy, not target accuracy.

We developed the NC objective set and $\delta=0.5$ on PACS source validation data without inspecting target outcomes, then evaluated them on 360 OfficeHome and TerraIncognita runs spanning five DG algorithms. On these post-development runs, AC-NC reduces mean target ECE, CwECE, and NLL relative to Source-Acc; the mean target-accuracy change is $+0.21$ percentage points. The $D_\infty$ aggregation was not fixed before these evaluations, so aggregation comparisons remain exploratory.

Our contributions are threefold:
\begin{itemize}
\item We characterize reliability variation among checkpoints with near-optimal source-validation accuracy within fixed training trajectories.
Differences in their target-domain probability quality reveal a reliability-selection opportunity not explicitly exploited by Source-Acc.

\item We propose accuracy-constrained reliability selection (AC), separating source-accuracy eligibility from reliability ranking. AC retains checkpoints within a prescribed tolerance of the best source-validation accuracy and ranks them by aggregating source reliability metrics normalized within the feasible set. It returns an existing checkpoint without target data, additional training, or weight averaging.

\item We conduct paired evaluations across three DG benchmarks and five training algorithms, with selection-rule, objective-set, aggregation, and accuracy-tolerance ablations. Post-development OfficeHome and TerraIncognita results show lower mean target ECE, CwECE, and NLL than Source-Acc, with small observed changes in mean target accuracy. Prediction-level case studies illustrate lower error rates above a fixed confidence threshold and, on subsets of shared errors, lower incorrect-prediction confidence and higher true-label probabilities.
\end{itemize}
\section{Related Work}
\paragraph{Model selection in domain generalization.}
DomainBed treats model selection as part of a complete DG algorithm and distinguishes training-domain from leave-one-domain-out validation \citep{gulrajani2021in}. \citet{lyu2023principled} filter candidates by validation loss and combine classification risk with feature-space domain discrepancy. Mixup-guided DG selection constructs a shifted validation set \citep{lu2023mixupselection}, while PAIR-s uses preference-aware ERM/OOD objective scoring and validation-accuracy filtering across runs \citep{chen2023pareto}. \citet[Section~5.1]{wald2021calibration} propose a directly related rule: select the model with lowest mean source-domain ECE subject to an in-domain validation-accuracy threshold. Their experimental procedure additionally recalibrates candidates before selection (Section~C.2 of their supplement). AC applies the accuracy-filtered selection principle directly to saved checkpoints within a fixed trajectory, without fitting a post-hoc calibrator. It expresses the threshold relative to the trajectory's best source accuracy; the reference rule ranks eligible checkpoints by an aggregate of NLL and CwECE, each min--max normalized within the feasible set. The selector uses source-validation predictive metrics, not algorithm-specific training-objective values or feature-space discrepancy estimates.
Our contribution is the fixed-trajectory analysis and evaluation of reliability-aware selection, rather than the accuracy-filtered selection principle itself.

\paragraph{Using training trajectories.}
SWAD averages weights densely over an overfit-aware interval of a training trajectory \citep{cha2021swad}. Ensemble of Averages (EoA) ensembles moving-average models from independent runs \citep{arpit2022ensemble}. AC instead selects one saved checkpoint, without weight averaging or prediction ensembling. Weight averaging also yields a single model for inference; AC's distinction is returning an existing checkpoint rather than constructing new weights. Our checkpoint-SWAD baseline is a sparse-checkpoint approximation of SWAD, with its implementation and BatchNorm handling described in Appendix~\ref{app:exp-baselines}.
\paragraph{Calibration under distribution shift.}
\citet{wald2021calibration} relate multi-domain calibration to invariant prediction under explicit assumptions. For a fixed classifier, scalar temperature scaling adjusts probabilities without changing class predictions \citep{guo2017calibration}, but calibration can deteriorate under distribution shift \citep{ovadia2019can}. \citet{gong2021confidence} develop temperature-scaling methods using multiple source calibration domains, without access to target-domain data at calibration time. AC instead chooses which saved classifier to deploy, without fitting a calibrator. This choice may change both probabilities and class predictions; its target-domain effects are evaluated empirically.
\section{Method}
\label{sec:method}

\begin{figure}[t]
\centering
\includegraphics[width=\linewidth]{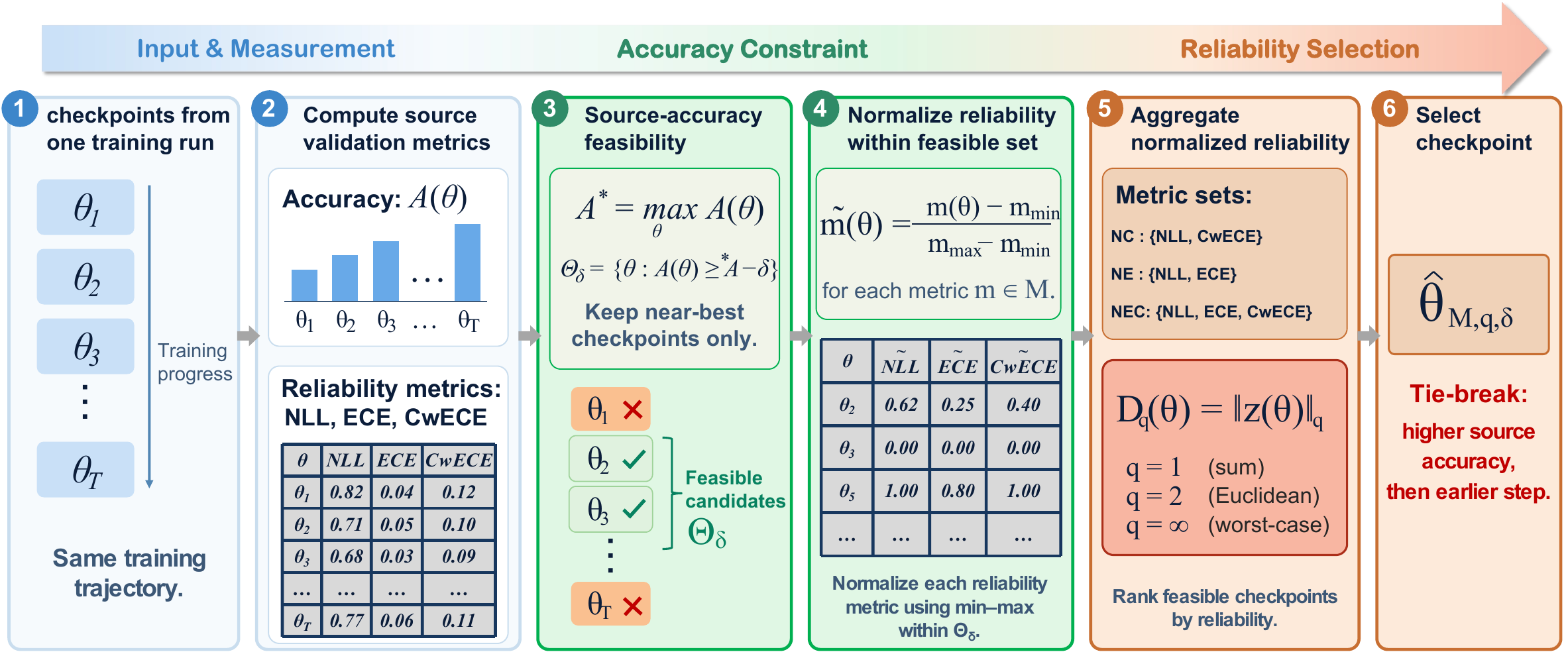}
\caption{Overview of the checkpoint selection process. The figure illustrates how checkpoints are evaluated based on source accuracy and reliability, and how the accuracy-constrained selection criterion is applied.}
\label{fig:method-overview}
\end{figure}

We study source-only checkpoint selection within a fixed training trajectory.
Source accuracy determines which checkpoints are eligible, and source
reliability ranks the eligible checkpoints. We first define the setting and a
fundamental limitation of source-only selection, then introduce
accuracy-constrained reliability selection (AC), and finally give a
finite-sample source-accuracy guarantee.
Figure~\ref{fig:method-overview} summarizes the selection pipeline.

\subsection{Problem setup and source-only limitation}
\label{sec:method-setup}

Let
\begin{equation}
\Theta=\{\theta_t:t\in\mathcal T\}
\end{equation}
be the finite, nonempty set of checkpoints saved during one DG training run,
with $T=|\mathcal T|$. Source domain $e\in\{1,\ldots,E\}$ has distribution
$P_e$ and held-out validation set $S_e$ of size $n_e\ge1$. Let $f_\theta$
denote a checkpoint's deterministic class prediction. We use equal-domain
source accuracy on the $[0,100]$ scale:
\begin{equation}
\begin{aligned}
A_{\mathrm{src}}(\theta)
&=
\frac{100}{E}
\sum_{e=1}^{E}
\Pr_{P_e}\!\left(f_\theta(X)=Y\right),\\
\widehat A_{\mathrm{src}}(\theta)
&=
\frac{100}{E}
\sum_{e=1}^{E}
\frac{1}{n_e}
\sum_{(x,y)\in S_e}
\mathbf 1\{f_\theta(x)=y\}.
\end{aligned}
\end{equation}
Source-Acc selects the earliest checkpoint maximizing
$\widehat A_{\mathrm{src}}$; denote it by $\theta_{\mathrm{SA}}$.
A source-only selector may use the saved trajectory and source-validation
statistics, but not target-domain observations.

\noindent\textbf{Proposition 1 (accuracy impossibility with unrestricted targets).}
Fix the source observations and a finite trajectory containing
$\theta_a,\theta_b$ such that
$f_{\theta_a}(x)\ne f_{\theta_b}(x)$ for some input $x$.
If target distributions are unrestricted, no source-only selector selects a
target-accuracy maximizer with probability one for every target domain.
For a randomized selector, the probability is over its internal randomization.

The proof constructs two target point masses with disjoint sets of
accuracy-maximizing checkpoints; see Appendix~\ref{app:prop1-proof}.
The result applies to Source-Acc and AC alike. It rules out a universal
target-accuracy guarantee without additional assumptions, but does not preclude
useful source-side selection under structured domain shifts.

\subsection{Accuracy-constrained reliability selection}
\label{sec:ac}

AC separates source-accuracy eligibility from reliability ranking. A
configuration specifies an accuracy tolerance $\delta\ge0$, a nonempty
reliability objective set $\mathcal M$, and an aggregation parameter
$q\in\{1,2,\infty\}$.

\paragraph{Accuracy eligibility.}
Define the empirical feasible set
\begin{equation}
\Theta_\delta
=
\left\{
\theta\in\Theta:
\widehat A_{\mathrm{src}}(\theta)
\ge
\widehat A_{\mathrm{src}}(\theta_{\mathrm{SA}})-\delta
\right\}.
\label{eq:accuracy-feasible-set}
\end{equation}
Because $\theta_{\mathrm{SA}}\in\Theta_\delta$, every AC selection
$\widehat\theta_{\mathrm{AC}}\in\Theta_\delta$ satisfies
\begin{equation}
0
\le
\widehat A_{\mathrm{src}}(\theta_{\mathrm{SA}})
-
\widehat A_{\mathrm{src}}(\widehat\theta_{\mathrm{AC}})
\le
\delta.
\end{equation}
Thus $\delta$ directly bounds the empirical source-accuracy deficit.

\paragraph{Reliability objectives.}
For reliability error $m$, let $m_{S_e}(\theta)$ denote its value on source
validation domain $e$, and average domains equally:
\begin{equation}
\widehat m_{\mathrm{src}}(\theta)
=
\frac{1}{E}
\sum_{e=1}^{E}m_{S_e}(\theta).
\end{equation}
Our reference objective set is
\begin{equation}
\mathcal M_{\mathrm{NC}}
=
\{\mathrm{NLL},\mathrm{CwECE}\}.
\label{eq:default-objectives}
\end{equation}
AC-ECE, AC-NLL, and AC-CwECE use singleton objective sets within the same accuracy-feasible set. AC-NC uses \(\mathcal M_{\mathrm{NC}}=\{\mathrm{NLL},\mathrm{CwECE}\}\) as a joint reference configuration.
NLL measures probability fit as a proper scoring rule \citep{gneiting2007strictly},
while CwECE measures class-wise calibration. The main experiments use Gaussian
soft-bin squared-gap variants of top-label ECE and class-wise ECE
\citep{guo2017calibration,kull2019beyond}; their
exact estimators and numerical conventions are given in
Appendix~\ref{app:calibration-metrics}. We additionally consider
$\mathcal M_{\mathrm{NE}}=\{\mathrm{NLL},\mathrm{ECE}\}$ and
$\mathcal M_{\mathrm{NEC}}
=\{\mathrm{NLL},\mathrm{ECE},\mathrm{CwECE}\}$
as objective-set ablations.

\paragraph{Within-set normalization.}
Reliability metrics have different scales, so AC normalizes each objective
within the same feasible set. For $m\in\mathcal M$, define
\begin{equation}
a_m
=
\min_{\theta\in\Theta_\delta}
\widehat m_{\mathrm{src}}(\theta),
\qquad
b_m
=
\max_{\theta\in\Theta_\delta}
\widehat m_{\mathrm{src}}(\theta),
\end{equation}
and, with $\eta=10^{-12}$,
\begin{equation}
\widetilde m(\theta)
=
\begin{cases}
0,
& b_m=a_m,\\[3pt]
\displaystyle
\frac{\widehat m_{\mathrm{src}}(\theta)-a_m}
     {b_m-a_m+\eta},
& b_m>a_m.
\end{cases}
\end{equation}
Thus each nonconstant objective is mapped approximately to $[0,1]$ over
$\Theta_\delta$. Numerical edge cases associated with this normalization are
discussed in Appendix~\ref{app:normalization-details}.

\paragraph{Aggregation and selection.}
Let
\begin{equation}
\widetilde{\mathbf L}^{\mathcal M}(\theta)
=
[\widetilde m(\theta)]_{m\in\mathcal M}.
\end{equation}
We aggregate normalized reliability errors by
\begin{equation}
D_q(\theta)
=
\left\|
\widetilde{\mathbf L}^{\mathcal M}(\theta)
\right\|_q
=
\begin{cases}
\left(
\sum_{m\in\mathcal M}\widetilde m(\theta)^q
\right)^{1/q},
&1\le q<\infty,\\[5pt]
\displaystyle
\max_{m\in\mathcal M}\widetilde m(\theta),
&q=\infty.
\end{cases}
\label{eq:reliability-distance}
\end{equation}
$D_\infty$ minimizes the largest normalized reliability error; $D_1$ and
$D_2$ give alternative aggregation rules. AC minimizes $D_q$ within
$\Theta_\delta$, breaking ties by higher source accuracy and then the earlier
training step:
\begin{equation}
\widehat\theta_{\mathrm{AC}}
=
\operatorname*{arg\,min}^{\mathrm{lex}}_{\theta_t\in\Theta_\delta}
\left(
D_q(\theta_t),
-\widehat A_{\mathrm{src}}(\theta_t),
t
\right).
\end{equation}
No target-domain quantity enters this selection rule. Since
$\theta_{\mathrm{SA}}\in\Theta_\delta$, AC cannot have a larger
feasible-set aggregate score than Source-Acc, but this does not imply
improvement in every constituent reliability metric or on the target domain.

\begin{algorithm}[tbp]
\caption{AC checkpoint selection}
\label{alg:ac-selection}
\begin{algorithmic}[1]
\REQUIRE Saved trajectory $\Theta$; per-domain source-validation accuracy and
reliability metrics; tolerance $\delta\ge0$ (pp); objective set $\mathcal M$;
$q\in\{1,2,\infty\}$; offset $\eta=10^{-12}$
\STATE Average validation metrics equally across source domains to obtain
$\widehat A_{\mathrm{src}}$ and $\widehat m_{\mathrm{src}}$,
$m\in\mathcal M$.
\STATE Compute
$A_{\max}=\max_{\theta\in\Theta}\widehat A_{\mathrm{src}}(\theta)$ and
$\Theta_\delta
=\{\theta:\widehat A_{\mathrm{src}}(\theta)\ge A_{\max}-\delta\}$.
\FOR{each $m\in\mathcal M$}
    \STATE Compute $a_m,b_m$ over $\Theta_\delta$ and normalize
    $\widehat m_{\mathrm{src}}$ to $\widetilde m$; use zero for a constant
    objective.
\ENDFOR
\STATE Compute $D_q(\theta)$ from
Eq.~\eqref{eq:reliability-distance} for $\theta\in\Theta_\delta$.
\STATE Minimize
$(D_q(\theta_t),-\widehat A_{\mathrm{src}}(\theta_t),t)$
lexicographically over $\Theta_\delta$.
\RETURN $\widehat\theta_{\mathrm{AC}}$.
\end{algorithmic}
\end{algorithm}

\subsection{Finite-sample source-accuracy control}
\label{sec:source-accuracy-bound}

The tolerance in Eq.~\eqref{eq:accuracy-feasible-set} directly controls
empirical source accuracy. Under standard sampling and validation-independence
conditions, Hoeffding's inequality \citep{hoeffding1963probability} also yields
a population source-accuracy bound.

\noindent\textbf{Proposition 2 (finite-sample source-accuracy bound).}
Condition on a trajectory $\Theta$ generated independently of the source
validation sets. Suppose $S_e$ contains $n_e$ i.i.d.\ samples from $P_e$,
independently across domains. For $\alpha\in(0,1)$, define
\begin{equation}
r_\alpha
=
100
\sqrt{
\frac{\log(2T/\alpha)}{2E^2}
\sum_{e=1}^{E}\frac{1}{n_e}
}.
\end{equation}
With probability at least $1-\alpha$, every
$\theta\in\Theta_\delta$ satisfies
\begin{equation}
A_{\mathrm{src}}(\theta)
\ge
\max_{\theta'\in\Theta}
A_{\mathrm{src}}(\theta')
-\delta-2r_\alpha.
\end{equation}

The event is uniform over the fixed candidate set, so the same validation
samples may subsequently be reused for reliability-based ranking.
The guarantee requires candidate generation to be independent of these
validation sets; validation feedback that changes the candidate trajectory is
not covered. It controls population source accuracy only, not reliability or
target-domain performance. The proof is given in
Appendix~\ref{app:source-bound-proof}.

More generally, source reliability ordering need not transfer to an unseen
target domain. Appendix~\ref{app:conditional-transfer} gives a conditional
population-level ordering result under a monotone source--target relation with
bounded checkpoint-dependent residual variation; the required target-dependent
quantities are unobserved and are not inputs to AC.

\section{Experiments}
\label{sec:experiments}
\suppressfloats[t]

\subsection{Experimental setup}
\label{sec:exp-setup}
We evaluate CORAL \citep{sun2016deepcoral}, ERM, GroupDRO
\citep{sagawa2020distributionally}, IRM \citep{arjovsky2019invariant}, and
VREx \citep{krueger2021rex} on PACS, OfficeHome, and TerraIncognita using
DomainBed source validation
\citep{gulrajani2021in,li2017deeper,venkateswara2017deep,beery2018recognition}.
Four held-out domains, three hyperparameter seeds, and three trial seeds give
$3\times5\times4\times3\times3=540$ runs, each with 51 checkpoints at steps
$0,100,\ldots,5000$. Selectors share each fixed trajectory; outcomes are
averaged equally across runs. Target evaluation uses the held-out domain's
\texttt{in} split (training details: Appendix~\ref{sec:training-settings}).

PACS source validation was used to develop NC and $\delta=0.5$, which were
then fixed for the 360 OfficeHome/TerraIncognita runs. The aggregation rule
was not fixed before these evaluations: distance comparisons remain
exploratory, and pooled results including PACS development are descriptive.

We report Gaussian soft-bin squared-gap ECE and CwECE
(Appendix~\ref{app:calibration-metrics}). Accuracy and calibration are scaled
by 100; NLL is unscaled. Paired changes are AC minus Source-Acc, so positive
accuracy and negative error changes favor AC. The 95\% percentile intervals
use 10,000 paired run-level bootstrap resamples \citep{efron1979bootstrap},
conditional on the observed trajectories; shared-seed dependence is not
modeled hierarchically (Appendix~\ref{app:exp-inference}). An accuracy
interval containing zero does not establish equivalence.

\subsection{Main results and heterogeneity}
\label{sec:exp-main}
\label{sec:exp-method-dataset}
On the 360 post-development runs, AC-NC/$D_\infty$ lowers mean target ECE,
CwECE, and NLL; all three paired intervals are below zero
(Table~\ref{tab:main}). The mean accuracy change is $+0.2133$ pp, but its
interval spans losses and gains. Thus these results support improved mean
probability quality, not target-accuracy preservation.

\begin{table}[htbp]
\centering
\caption{Target results for AC-NC ($D_\infty$, $\delta=0.5$).
Top: dataset means, 180 runs each; Gap is the source-accuracy deficit (pp).
Bottom: AC-NC minus Source-Acc on the 360 post-development runs, with 95\%
paired intervals. Accuracy differences are in pp, calibration is scaled by
100, and NLL is unscaled. PACS is development; distance choice is exploratory.}
\label{tab:main}
\label{tab:pairwise}
\small
\begin{tabular*}{\linewidth}{@{\extracolsep{\fill}}llrrrrr@{}}
\toprule
Dataset & Selector & Gap & Acc. $\uparrow$ & ECE $\downarrow$ & CwECE $\downarrow$ & NLL $\downarrow$\\
\midrule
OfficeHome & Source-Acc & 0.0000 & 60.87 & 2.77 & 3.32 & 4.2724\\
& AC-NC & 0.1612 & 61.06 & 2.48 & 3.05 & 4.2362\\
TerraInc. & Source-Acc & 0.0000 & 42.74 & 9.10 & 11.35 & 2.3438\\
& AC-NC & 0.1091 & 42.97 & 8.90 & 11.25 & 2.3210\\
\midrule
PACS (dev.) & Source-Acc & 0.0000 & 80.73 & 1.63 & 2.75 & 0.7794\\
& AC-NC & 0.1647 & 80.55 & 1.59 & 2.75 & 0.7728\\
\bottomrule
\end{tabular*}
\par
\begin{tabular*}{\linewidth}{@{\extracolsep{\fill}}lrrrr@{}}
\toprule
360 runs & $\Delta$ Acc. & $\Delta$ ECE & $\Delta$ CwECE & $\Delta$ NLL\\
\midrule
Mean & $+0.2133$ & $-0.2395$ & $-0.1821$ & $-0.0295$\\
95\% CI & $[-0.0592,+0.5017]$ & $[-0.4038,-0.0903]$ & $[-0.2841,-0.0803]$ & $[-0.0483,-0.0119]$\\
\bottomrule
\end{tabular*}
\end{table}

Gains are heterogeneous: mean CwECE and NLL both decrease for all five
algorithms on OfficeHome, three on TerraIncognita, and two on PACS
(Appendix Table~\ref{tab:dataset-method-deltas}). Across the pooled 540 runs,
118 selections lose target accuracy, including 76 (14.1\%) that lose at
least one pp; the fifth percentile is $-3.0$ pp
(Appendix~\ref{app:accuracy-losses}). Pooled objective-set contrasts
are in Appendix Table~\ref{tab:six-variant}.
Figure~\ref{fig:ac-nc-across-methods} shows mostly lower target NLL across
algorithm--selector combinations, with variable accuracy changes and
source--target rank agreement.

\subsection{Prediction-level diagnostics}
\label{sec:prediction-cases}
\begin{figure}[H]
\centering
\includegraphics[width=\linewidth]{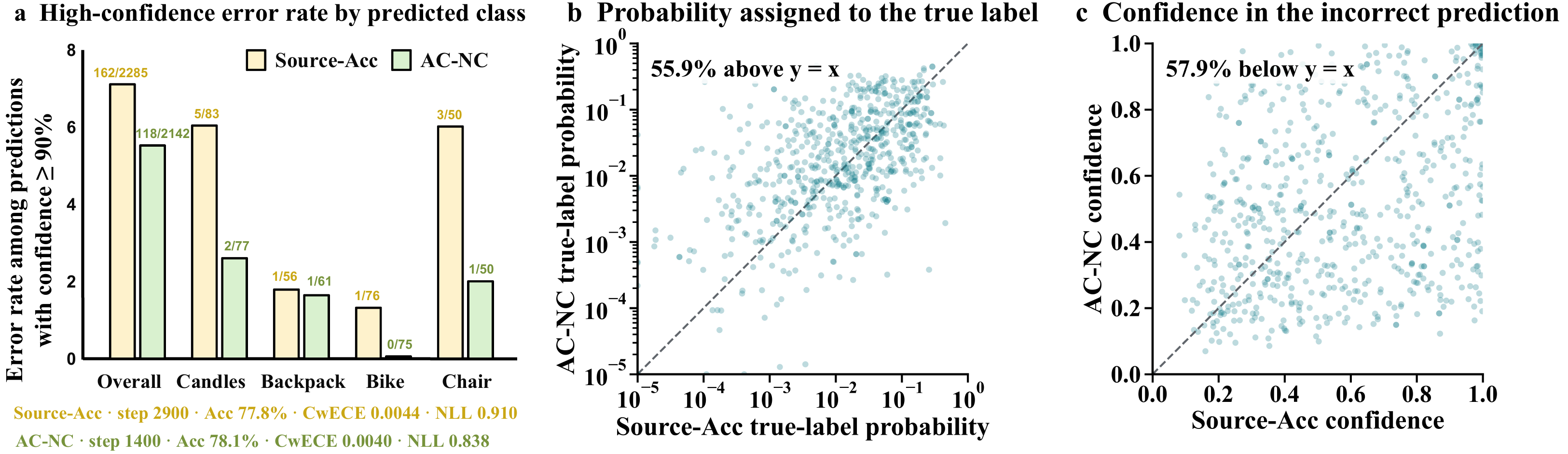}
\caption{Post-hoc target diagnostics. (a) Errors among predictions with
confidence $\ge90\%$; labels show errors/high-confidence predictions.
(b,c) OfficeHome Art/ERM: true-label probability and incorrect-prediction
confidence on 642 shared errors; dashed lines indicate equality. Samples
use the aligned target \texttt{in} split; CwECE in (a) is unscaled.}
\label{fig:prediction-cases}
\end{figure}

Figure~\ref{fig:prediction-cases}(a) shows a post-hoc case in which the error
rate above 90\% confidence falls from 7.09\% to 5.51\%, but coverage also
falls; this is not an equal-coverage comparison. In panels (b,c), among 642
shared errors, AC-NC increases true-label probability in 55.9\% and reduces
incorrect-prediction confidence in 57.9\%. These cases illustrate
probability shifts rather than establish general prediction-level
improvement (additional images: Appendix~\ref{sec:qualitative-cases}).

\subsection{What drives checkpoint reselection?}
\label{sec:exp-feasible}
\label{sec:exp-aggregation}
\paragraph{Accuracy filtering.}
On OfficeHome, unconstrained ECE/CwECE selection chooses step 0 in
94.4\%/95.6\% of runs, with target accuracy of 2.55\%/1.91\%. Pure NLL
avoids this collapse but has no explicit accuracy budget
(Appendix Table~\ref{tab:pure-collapse}). At $\delta=0.5$, 391/540 runs have
multiple eligible checkpoints, and AC-NC differs from Source-Acc in 264 of
these (67.5\%). Reliability still varies within the feasible sets
(Appendix Table~\ref{tab:plateau-variation}).

\paragraph{Joint versus single-objective ranking.}
AC-ECE, AC-NLL, and AC-CwECE change only the ranking objective.
Table~\ref{tab:component-main} reports pooled paired comparisons with
AC-NLL and AC-CwECE: all target-metric intervals include zero, establishing
neither superiority nor equivalence of the joint rule. Table~\ref{tab:component-pacs} retains the PACS-only comparison, including
AC-ECE and three feasible-set controls. AC-early uses the earliest eligible
checkpoint; AC-random uses one uniform draw per run (seed 20260926);
AC-mean reports the expected metric under uniform checkpoint sampling,
not weight averaging or $D_1$ score aggregation. Although AC-NC has the
best means among the four reliability-ranking rules on PACS, these are
development results, not evidence of post-development superiority.

\begin{table}[H]
\centering
\caption{Joint versus single-objective ranking on 540 paired runs, including
PACS development (descriptive). Entries are AC-NC minus the comparator,
with 95\% paired intervals. All selectors use the same $\Theta_{0.5}$.
Accuracy differences are in pp; calibration is scaled by 100. Full
precision and secondary outcomes: Appendix Table~\ref{tab:component-ci}.}
\label{tab:component-main}
\small
\setlength{\tabcolsep}{3pt}
\begin{tabular*}{\linewidth}{@{\extracolsep{\fill}}lrrrr@{}}
\toprule
Comparator & $\Delta$ Acc. & $\Delta$ ECE & $\Delta$ CwECE & $\Delta$ NLL\\
\midrule
AC-NLL & $+0.0223$ & $-0.0240$ & $-0.0424$ & $-0.0003$\\
& $[-0.1061,+0.1458]$ & $[-0.0900,+0.0460]$ & $[-0.0948,+0.0097]$ & $[-0.0084,+0.0083]$\\
AC-CwECE & $-0.0423$ & $+0.0092$ & $+0.0084$ & $+0.0013$\\
& $[-0.1853,+0.1029]$ & $[-0.0942,+0.1303]$ & $[-0.0590,+0.0753]$ & $[-0.0118,+0.0164]$\\
\bottomrule
\end{tabular*}
\end{table}

\begin{table}[htbp]
\centering
\caption{Ranking objectives and feasible-set controls on PACS development
only (180 runs, $\delta=0.5$). Gap is the source-accuracy deficit (pp);
accuracy and calibration are scaled by 100. These descriptive means do
not establish superiority of joint ranking.}
\label{tab:component-pacs}
\small
\begin{tabular*}{\linewidth}{@{\extracolsep{\fill}}lrrrrr@{}}
\toprule
Selector & Gap & Acc. $\uparrow$ & ECE $\downarrow$ & CwECE $\downarrow$ & NLL $\downarrow$\\
\midrule
Source-Acc & 0.0000 & 80.7253 & 1.6265 & 2.7526 & 0.7794\\
AC-ECE & 0.2167 & 80.2406 & 1.5928 & 2.8638 & 0.7835\\
AC-NLL & 0.1689 & 80.3712 & 1.6402 & 2.8355 & 0.7787\\
AC-CwECE & 0.1660 & 80.5422 & 1.6172 & 2.8025 & 0.7801\\
AC-NC & 0.1647 & 80.5523 & \textbf{1.5924} & \textbf{2.7501} & \textbf{0.7728}\\
AC-mean & 0.1933 & 80.3927 & 1.6477 & 2.8624 & 0.8020\\
AC-early & 0.2220 & 79.9796 & 1.5933 & 2.8752 & 0.8025\\
AC-random & 0.1781 & 80.3966 & 1.6895 & 2.8563 & 0.8078\\
\bottomrule
\end{tabular*}
\end{table}

\begin{figure}[H]
\centering
\includegraphics[width=\linewidth]{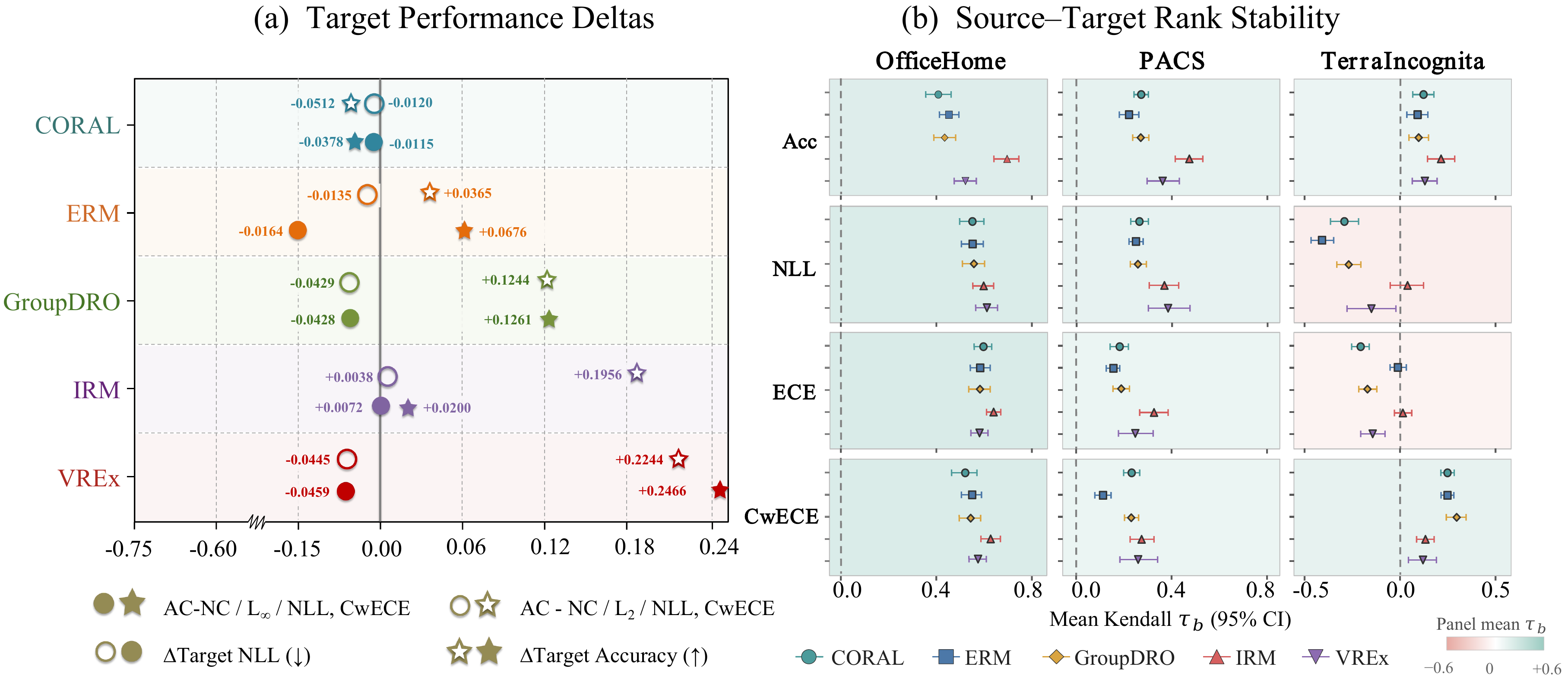}
\caption{
Cross-algorithm and cross-dataset evaluation of accuracy-constrained
reliability selection. (a) Mean paired changes in target accuracy (stars,
percentage points) and NLL (circles, original scale) relative to Source-Acc
for five DG training algorithms, pooled over OfficeHome, PACS, and
TerraIncognita. Open markers denote AC-NC/$D_2$ and filled markers denote
AC-NEC/$D_\infty$. (b) Mean source--target Kendall $\tau_b$ of checkpoint
rankings for accuracy, NLL, ECE, and CwECE across the same algorithms and
datasets; bars show 95\% confidence intervals and shading shows the panel
mean. Target NLL decreases for most algorithm--selector combinations,
although the effects and rank agreement vary across datasets.
}
\label{fig:ac-nc-across-methods}
\end{figure}

\paragraph{Aggregation and objective sets.}
With NC fixed, $D_1$, $D_2$, and $D_\infty$ give similar aggregate outcomes;
$D_1$ and $D_\infty$ disagree in only 19/540 selections. At fixed
$D_\infty$, NEC's larger pooled NLL reduction is sensitive to one
OfficeHome/IRM run. These comparisons establish neither a preferred distance
nor a benefit from adding ECE to NC (Appendices~\ref{sec:validation-results}
and~\ref{app:aggregation-details}).

\subsection{Tolerance and additional comparisons}
\label{sec:exp-tolerance}
\label{sec:exp-temperature-scaling}
\paragraph{Tolerance sensitivity.}
Increasing $\delta$ expands the candidate set and changes its normalization.
At $\delta=1.0$, pooled reliability errors are lower than at $\delta=0.5$
(Table~\ref{tab:delta-nc}); however, paired accuracy intervals for both
alternative tolerances include zero, as does the NLL interval for
$\delta=1.0$. Dataset-level patterns identify no uniformly preferable
value (Appendix~\ref{sec:tolerance-ablation}).

\begin{table}[htbp]
\centering
\caption{Tolerance sensitivity on 540 pooled runs (including PACS
development). Candidates is the mean feasible-set size; Gap is the mean
source-accuracy deficit (pp). Accuracy and calibration are scaled by 100.}
\label{tab:delta-nc}
\small
\begin{tabular*}{\linewidth}{@{\extracolsep{\fill}}lrrrrrr@{}}
\toprule
Selector & Candidates & Gap & Acc. $\uparrow$ & ECE $\downarrow$ & CwECE $\downarrow$ & NLL $\downarrow$\\
\midrule
Source-Acc & -- & 0.0000 & 61.44 & 4.50 & 5.81 & 2.4652\\
AC-NC ($\delta=0.1$) & 1.30 & 0.0063 & 61.46 & 4.46 & 5.78 & 2.4610\\
AC-NC ($\delta=0.5$) & 3.92 & 0.1450 & 61.53 & 4.33 & 5.69 & 2.4433\\
AC-NC ($\delta=1.0$) & 9.71 & 0.3516 & \textbf{61.51} & \textbf{4.19} & \textbf{5.62} & \textbf{2.1109}\\
\bottomrule
\end{tabular*}
\end{table}

\paragraph{Averaging and validation protocol.}
In an independent 180-run OfficeHome rerun, checkpoint-SWAD (a sparse
approximation) with BatchNorm recalibration exceeds AC-NC in mean accuracy
(62.62\% versus 60.97\%) and lowers all three reliability errors
(Appendix~\ref{app:exp-baselines}). In the separate ERM/VREx LODO study,
LODO-Acc has lower NLL and ECE than ordinary AC. Within LODO, reliability
ranking reduces fixed-hyperparameter ECE by $0.562$
(95\% CI for AC-LODO minus LODO-Acc: $[-1.160,-0.099]$), without a
resolved accuracy difference. LODO uses distinct hard-bin calibration
metrics and additional training; these cohorts are not pooled with
Table~\ref{tab:main} (Appendix~\ref{app:lodo-details}).

\paragraph{Temperature scaling (TS).}
A separate study fits one scalar temperature by minimizing NLL on pooled
source \texttt{out} samples after checkpoint selection; targets are used
only for evaluation. TS leaves accuracy unchanged. After TS, the selectors
have similar ECE, while AC-NC has slightly higher CwECE and NLL; ECE
improves in only 5/10 dataset--algorithm cells for either selector. Thus the
operations can be combined, but gains are not consistently additive
(Table~\ref{tab:temperature-scaling-main}).

\begin{table}[htbp]
\centering
\caption{Separate source-fitted TS study on OfficeHome/TerraIncognita.
Entries equally average 10 dataset--algorithm cells (36 matched runs each);
arrows show raw $\to$ TS values. Calibration is scaled by 100.}
\label{tab:temperature-scaling-main}
\small
\begin{tabular*}{\linewidth}{@{\extracolsep{\fill}}lcccc@{}}
\toprule
Selector & Acc. (\%) $\uparrow$ & ECE $\downarrow$ & CwECE $\downarrow$ & NLL $\downarrow$\\
\midrule
Source-Acc & 51.94 & $5.578\to5.003$ & $7.247\to6.468$ & $2.183\to1.950$\\
AC-NC & 51.99 & $5.481\to4.993$ & $7.155\to6.497$ & $2.190\to1.969$\\
\bottomrule
\end{tabular*}
\end{table}
\FloatBarrier

\section{Conclusion}

In the evaluated DG trajectories, checkpoints with similar source-validation accuracy can differ in probability quality. Reliability-aware reselection improves mean target probability quality relative to Source-Acc on the original post-development cohort. AC provides a concrete selection protocol, but the current evidence does not establish superiority of its joint NLL--CwECE reference over accuracy-constrained single-objective ranking.

\section*{AI Use Statement}
AI tools were used solely to assist with writing and language polishing, including improving clarity and readability, and to support literature retrieval and the discovery of relevant prior work.
The authors take full responsibility for the content of this paper, including the accuracy of its claims, results, and references.

\bibliographystyle{iclr2027_conference}
\bibliography{references}

\clearpage

\appendix
\numberwithin{equation}{section}
\numberwithin{table}{section}
\numberwithin{figure}{section}

\section{Qualitative Checkpoint Case Studies}
\label{sec:qualitative-cases}

\begin{figure}[H]
\centering
\includegraphics[width=\linewidth,trim=45bp 195bp 215bp 137bp,clip]{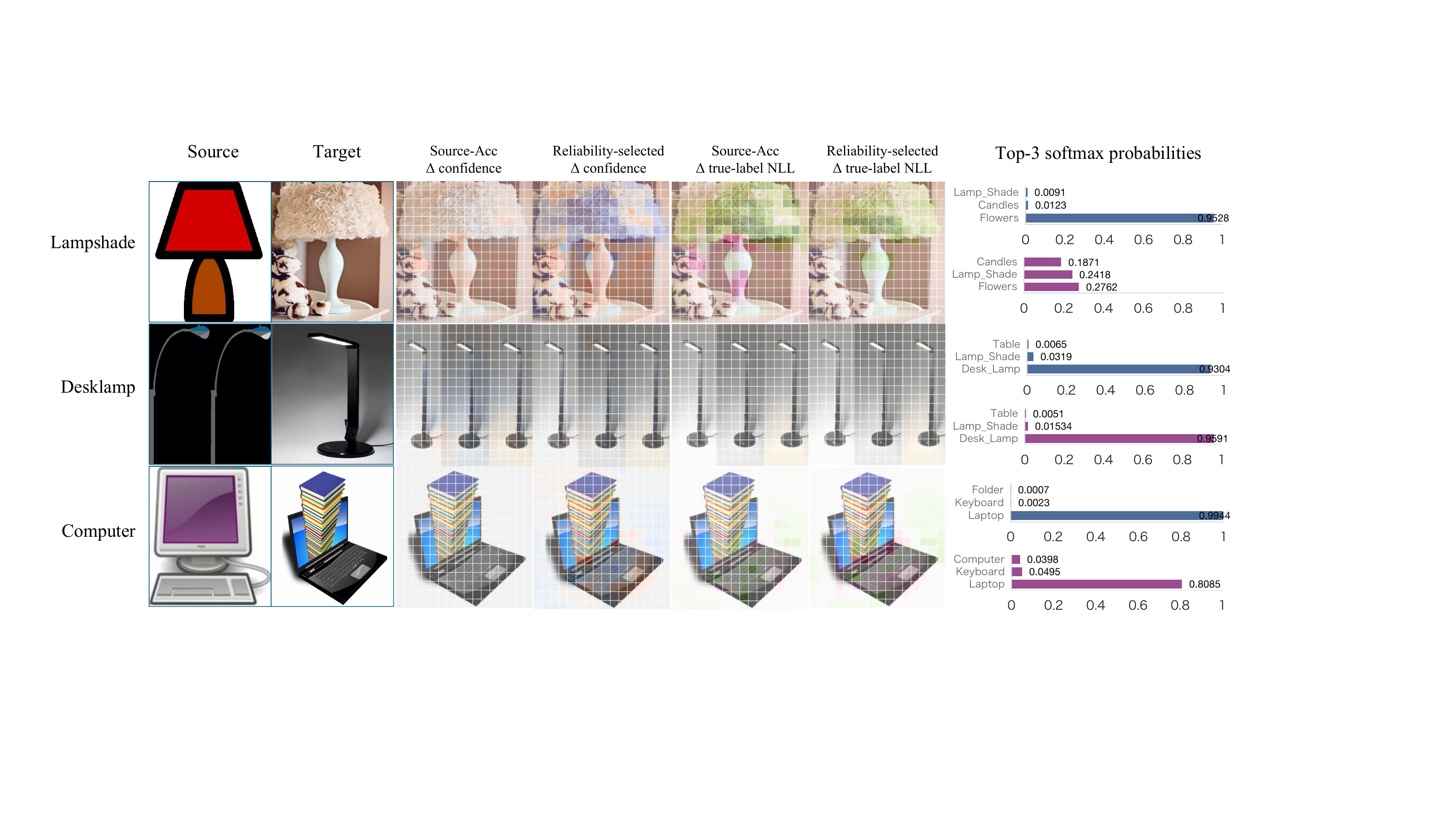}
\caption{Target-domain examples comparing Source-Acc and reliability-selected checkpoints. The left columns show a source-domain class exemplar and a target image. Each heat map masks one $16\times16$ patch: $\Delta$ confidence is the decrease in top-1 probability (purple: support for the prediction; blue: counter-evidence), whereas $\Delta$ true-label NLL is the increase in true-label loss (green: support for the true class; magenta: conflicting evidence). Colors are comparable within a metric, not between metrics. Right-hand bars show the three largest softmax probabilities.}
\label{fig:checkpoint-qualitative}
\end{figure}

\noindent\textbf{Lampshade.} Both checkpoints misclassify the target image as Flowers. Reliability selection reduces the Flowers probability from $0.9528$ to $0.2762$ while increasing the true-class Lamp\_Shade probability from $0.0091$ to $0.2418$. The top-1 error remains, but its confidence is substantially lower and more probability mass is assigned to the correct class.

\noindent\textbf{Desklamp.} Both checkpoints predict Desk\_Lamp correctly; its probability rises from $0.9304$ to $0.9591$. For the confidence bin containing this example, the empirical confidence--accuracy gap decreases from $0.226$ to $0.001$, while example-level NLL decreases from $0.07$ to $0.04$. The diffuse occlusion responses indicate that no single $16\times16$ patch dominates the prediction.

\noindent\textbf{Computer.} Both checkpoints incorrectly predict Laptop, but its probability falls from $0.9944$ to $0.8085$ after reliability selection. The true-class Computer probability rises from approximately $0.0002$ to $0.0398$, reducing example-level NLL from $8.53$ to $3.22$. Green regions near the screen and keyboard indicate local support for the true class, although that evidence does not change the top-1 decision.

These examples illustrate lower confidence in shared errors and greater retention of true-class probability without implying that every misclassification is corrected. Aggregate calibration and NLL outcomes are evaluated separately in the main paper.

\clearpage

\section{Reliability Metrics and Selection Details}
\label{app:reliability-details}

This section fixes the metric and normalization conventions used throughout
the supplement. We first define the Gaussian soft-bin reliability metrics for
the main checkpoint-selection and checkpoint-SWAD cohorts, then distinguish
the hard-bin metrics used only in the LODO study, and finally specify the
within-feasible-set normalization used by AC.

\subsection{Main-cohort reliability metrics}
\label{app:calibration-metrics}
\label{sec:calibration-definitions}

This section gives the exact reliability estimators used by the main
checkpoint-selection experiments.

Let
\[
\mathcal V=\{(x_i,y_i)\}_{i=1}^{n}
\]
be a validation set with $C$ classes. Let $z_i$ denote the logits,
\[
p_{ic}=\operatorname{softmax}(z_i)_c,
\qquad
u_i=\max_c p_{ic},
\]
and
\[
a_i
=
\mathbf 1\{\arg\max_c p_{ic}=y_i\}.
\]

\paragraph{Negative log-likelihood.}
We compute
\[
\mathrm{NLL}_{\mathcal V}
=
-\frac{1}{n}
\sum_{i=1}^{n}
\log p_{i,y_i},
\]
implemented using stable logit-based cross-entropy.

\paragraph{Gaussian soft-bin calibration scores.}
The main experiments use Gaussian soft-bin, squared-gap calibration errors.
Let $\mu_b\in[0,1]$, $b=1,\ldots,B$, be equally spaced bin centers and let
$h>0$ be the run-fixed bandwidth. For top-label calibration, define
\begin{equation}
w_{ib}
=
\frac{
\exp[-(u_i-\mu_b)^2/(2h^2)]
}{
\sum_{r=1}^{B}
\exp[-(u_i-\mu_r)^2/(2h^2)]
}.
\label{eq:app-toplabel-weights}
\end{equation}
For class-wise calibration, define
\begin{equation}
v_{icb}
=
p_{ic}
\exp[-(p_{ic}-\mu_b)^2/(2h^2)].
\label{eq:app-classwise-weights}
\end{equation}
Unlike $w_{ib}$, $v_{icb}$ is not normalized across bins for each example and
contains an additional factor $p_{ic}$.

Let
\[
W_b=\sum_i w_{ib},
\qquad
V_{cb}=\sum_i v_{icb}.
\]
For positive denominators, define
\[
\begin{aligned}
\bar a_b
&=
\frac{\sum_i w_{ib}a_i}{W_b},
&
\bar u_b
&=
\frac{\sum_i w_{ib}u_i}{W_b},
\\
\bar y_{cb}
&=
\frac{
\sum_i v_{icb}\mathbf 1\{y_i=c\}
}{V_{cb}},
&
\bar p_{cb}
&=
\frac{\sum_i v_{icb}p_{ic}}{V_{cb}}.
\end{aligned}
\]

The resulting squared-gap calibration scores are
\begin{equation}
\mathrm{ECE}_{\mathcal V}
=
\sum_{b=1}^{B}
\frac{W_b}{n}
(\bar a_b-\bar u_b)^2
\label{eq:app-ece}
\end{equation}
\label{eq:ece-soft-definition}
and
\begin{equation}
\mathrm{CwECE}_{\mathcal V}
=
\frac{1}{C}
\sum_{c=1}^{C}
\sum_{b=1}^{B}
\frac{V_{cb}}{\sum_r V_{cr}}
(\bar y_{cb}-\bar p_{cb})^2.
\label{eq:app-cwece}
\end{equation}
\label{eq:cwece-soft-definition}
Class-wise terms use all $n$ validation examples, and the outer average weights
the $C$ classes equally. Empty-weight terms contribute zero.

All source reliability metrics are computed separately within each source
validation domain and then averaged equally:
\begin{equation}
\widehat m_{\mathrm{src}}(\theta)
=
\frac{1}{E}
\sum_{e=1}^{E}
m_{S_e}(\theta),
\qquad
m\in\mathcal M.
\label{eq:app-domain-average}
\end{equation}

\subsection{Implementation conventions}

The Gaussian soft-bin metrics use equally spaced centers
$\mu_b=(b-1)/(B-1)$ and a run-fixed bandwidth $h$. Each run uses the same
$(B,h)$ for all checkpoints, and the selector consumes the logged scores
without rebinning. The local exporter clamps denominators below $10^{-8}$ and
defaults to $B=15$ and $h=0.1$ unless run-specific hyperparameters override
them; the checkpoint-SWAD rerun uses 15 centers.

\subsection{LODO hard-bin calibration metrics}
\label{app:lodo-calibration-metrics}

The LODO analyses in Tables~\ref{tab:lodo-fixed-detailed}
and~\ref{tab:lodo-combined-contrasts} use a separate hard-bin, absolute-gap
definition.
Partition $[0,1]$ into $B$ equal-width intervals $I_b=((b-1)/B,b/B]$, including zero in the first interval. Following \citet{guo2017calibration}, define the top-label bins $B_b=\{i:u_i\in I_b\}$, with
\[
\operatorname{acc}(B_b)=\frac{1}{|B_b|}\sum_{i\in B_b}a_i,
\qquad
\operatorname{conf}(B_b)=\frac{1}{|B_b|}\sum_{i\in B_b}u_i.
\]
The expected calibration error is
\begin{equation}
\mathrm{ECE}=\sum_{b=1}^{B}\frac{|B_b|}{n}
\left|\operatorname{acc}(B_b)-\operatorname{conf}(B_b)\right|.
\label{eq:ece-definition}
\end{equation}

For class $c$, form the one-vs-rest bins $B_{c,b}=\{i:p_{ic}\in I_b\}$ over all $n$ examples and define
\[
\operatorname{freq}_c(B_{c,b})=\frac{1}{|B_{c,b}|}\sum_{i\in B_{c,b}}\mathbf 1\{y_i=c\},
\qquad
\operatorname{conf}_c(B_{c,b})=\frac{1}{|B_{c,b}|}\sum_{i\in B_{c,b}}p_{ic}.
\]
Let $n_c=\sum_{i=1}^{n}\mathbf 1\{y_i=c\}$ be the number of examples with true label $c$. We define class-weighted ECE as
\begin{equation}
\mathrm{CwECE}=\sum_{c=1}^{C}\frac{n_c}{n}
\sum_{b=1}^{B}\frac{|B_{c,b}|}{n}
\left|\operatorname{freq}_c(B_{c,b})-\operatorname{conf}_c(B_{c,b})\right|.
\label{eq:cwece-definition}
\end{equation}
Here, $|B_{c,b}|/n$ weights bins within each one-vs-rest error, and $n_c/n$ weights classes by their sample proportions. Empty bins and classes with $n_c=0$ contribute zero. These absolute-gap, frequency-weighted hard-bin metrics are distinct from the squared-gap, uniformly averaged soft-bin metrics above; results are compared only within their respective panels.

\subsection{Feasible-set normalization}
\label{app:normalization-details}

AC normalizes every reliability objective using extrema computed only over the
current accuracy-feasible set $\Theta_\delta$. Consequently, changing
$\delta$ affects both the candidate set and the scale on which candidates are
ranked.

For objective $m$, recall
\[
a_m
=
\min_{\theta\in\Theta_\delta}
\widehat m_{\mathrm{src}}(\theta),
\qquad
b_m
=
\max_{\theta\in\Theta_\delta}
\widehat m_{\mathrm{src}}(\theta),
\]
and
\[
\widetilde m(\theta)
=
\begin{cases}
0,&b_m=a_m,\\[3pt]
\displaystyle
\frac{\widehat m_{\mathrm{src}}(\theta)-a_m}
{b_m-a_m+\eta},
&b_m>a_m,
\end{cases}
\qquad
\eta=10^{-12}.
\]
Constant objectives therefore contribute zero.

Min--max normalization can amplify small raw differences when the range
$b_m-a_m$ is small. A simple edge case also shows why the numerical offset is
formally part of the specified rule. Suppose exactly two checkpoints are
eligible and their NLL and CwECE rankings are opposed. With exact min--max
normalization and no offset, their two-objective vectors are
\[
(0,1)
\qquad\text{and}\qquad
(1,0).
\]
They therefore tie under $D_1$, $D_2$, and $D_\infty$. The nonzero offset can
break this otherwise exact symmetry through the raw objective ranges. We
therefore treat $\eta$ as part of the implementation specification rather than
as a mathematically inert numerical convention.

\section{Per-Dataset and Per-Algorithm Results}
\label{app:dataset-algorithm-results}

Aggregate means can conceal dataset- and algorithm-specific failures. We
therefore report this breakdown immediately after the metric definitions.
PACS remains the development benchmark, whereas OfficeHome and
TerraIncognita constitute the post-development evaluation cohort.
Table~\ref{tab:dataset-method-deltas} resolves the aggregate AC-NC result by
dataset and training algorithm.

\begin{table}[H]
\centering
\caption{Default AC-NC minus Source-Acc for each dataset--algorithm group (36 matched runs per row). Gap is the Source-Acc minus AC-NC source-accuracy difference. Accuracy and worst held-out-domain (WHD) deltas are in percentage points; calibration deltas are multiplied by 100, and NLL retains its original scale.}
\label{tab:dataset-method-deltas}
\resizebox{\linewidth}{!}{%
\begin{tabular}{llrrrrrr}
\toprule
Dataset & Algorithm & Gap (pp) & $\Delta$ target acc. & $\Delta$ ECE & $\Delta$ CwECE & $\Delta$ NLL & $\Delta$ WHD \\
\midrule
OfficeHome & CORAL    & 0.2159 & $+0.13$ & $-0.39$ & $-0.33$ & $-0.0468$ & $-0.09$ \\
           & ERM      & 0.1798 & $+0.35$ & $-0.29$ & $-0.32$ & $-0.0438$ & $-0.05$ \\
           & GroupDRO & 0.2264 & $+0.20$ & $-0.45$ & $-0.39$ & $-0.0527$ & $+0.00$ \\
           & IRM      & 0.0449 & $+0.09$ & $-0.07$ & $-0.09$ & $-0.0074$ & $+0.00$ \\
           & VREx     & 0.1391 & $+0.19$ & $-0.23$ & $-0.22$ & $-0.0302$ & $-0.12$ \\
\midrule
PACS       & CORAL    & 0.1677 & $-0.31$ & $-0.06$ & $+0.09$ & $+0.0000$ & $+0.97$ \\
           & ERM      & 0.2240 & $-0.69$ & $+0.08$ & $+0.20$ & $+0.0192$ & $-1.58$ \\
           & GroupDRO & 0.1814 & $-0.09$ & $-0.08$ & $-0.08$ & $-0.0152$ & $-0.04$ \\
           & IRM      & 0.0995 & $-0.30$ & $+0.02$ & $+0.01$ & $+0.0059$ & $-0.97$ \\
           & VREx     & 0.1509 & $+0.53$ & $-0.14$ & $-0.18$ & $-0.0432$ & $+1.26$ \\
\midrule
TerraInc.  & CORAL    & 0.0881 & $+0.08$ & $+0.14$ & $+0.09$ & $+0.0122$ & $-0.38$ \\
           & ERM      & 0.2155 & $+0.53$ & $-0.28$ & $-0.40$ & $-0.0246$ & $+0.64$ \\
           & GroupDRO & 0.0846 & $+0.26$ & $-0.56$ & $-0.14$ & $-0.0604$ & $+0.28$ \\
           & IRM      & 0.0776 & $+0.28$ & $+0.22$ & $+0.24$ & $+0.0230$ & $+2.16$ \\
           & VREx     & 0.0798 & $+0.03$ & $-0.47$ & $-0.25$ & $-0.0644$ & $+0.84$ \\
\bottomrule
\end{tabular}
}
\end{table}
\FloatBarrier

\subsection{Distribution of target-accuracy changes}
\label{app:accuracy-losses}

The pooled 540-run analysis includes PACS development and is descriptive.
Although the mean NC/$D_\infty$ target-accuracy change is small and positive,
run-level effects are heterogeneous. Relative to Source-Acc, AC-NC loses target
accuracy in 118/540 runs. In 76/540 runs (14.1\%), the loss is at least one
percentage point, and the empirical fifth percentile of the target-accuracy
change is $-3.0$ percentage points.

These statistics emphasize that a source-side tolerance does not bound the
target-accuracy change and that an improvement in the pooled mean does not
imply per-run preservation.

\section{Additional Baselines and Validation Protocols}
\label{app:additional-baselines}

This section compares AC with alternatives that change either the deployed
model or the validation signal. Checkpoint-SWAD constructs new weights,
LODO uses additional leave-domain-out training runs, and the PAIR-s-style
diagnostic is restricted to VREx. Because these protocols use different cohorts
or model-construction procedures, each comparison states its own scope and
should not be treated as one pooled ranking.

\subsection{Checkpoint-SWAD on OfficeHome}
\label{app:exp-baselines}

\paragraph{Checkpoint-SWAD protocol.}
We evaluate checkpoint-SWAD in an independent, design-matched OfficeHome rerun.
The cohort contains five algorithms, four held-out domains, three
hyperparameter seeds, and three trial seeds, giving 180 runs.
Because the retained trajectories contain one endpoint every 100 training
steps, this experiment is a sparse-checkpoint approximation to SWAD rather than
an exact reproduction of dense online averaging. We evaluate all 51 endpoints
on the source-validation domains, average source NLL across domains, and replay
the LossValley queue with $n_{\mathrm{converge}}=3$,
$n_{\mathrm{tolerance}}=6$, and tolerance ratio $0.3$. The parameters of the
selected endpoints are averaged with the multiplicities induced by the queue.
No target-domain observation enters valley construction, parameter averaging,
or BatchNorm processing.

The training trajectories use unfrozen BatchNorm. For the primary
checkpoint-SWAD variant, we therefore reset the running statistics of the
averaged model and update them using 500 minibatches from the source training
domains. A paired ablation instead restores all non-parameter buffers from the
first selected endpoint and performs no BatchNorm update. Both variants use the
same selected endpoints, averaged parameters, and target examples.

\begin{table}[ht]
\centering
\caption{Design-matched checkpoint-SWAD results on OfficeHome (180 runs). Accuracy is in percent; Gaussian soft-bin squared-gap ECE and CwECE are multiplied by 100; NLL is unscaled. The two checkpoint-SWAD variants use identical averaged parameters and differ only in BatchNorm handling.}
\label{tab:swad-detailed}
\small
\setlength{\tabcolsep}{4pt}
\begin{tabular}{lrrrr}
\toprule
Selector & Acc. $\uparrow$ & ECE $\downarrow$ & CwECE $\downarrow$ & NLL $\downarrow$ \\
\midrule
Source-Acc & 60.85 & 2.61 & 3.26 & 2.1211 \\
AC-NC & 60.97 & 2.36 & 3.05 & 2.1156 \\
checkpoint-SWAD + BN recal. & 62.62 & 1.95 & 2.91 & 2.0026 \\
checkpoint-SWAD, no BN recal. & 48.36 & 11.99 & 3.35 & 9567.3991 \\
\bottomrule
\end{tabular}
\end{table}

\paragraph{Comparison with checkpoint selection.}
On OfficeHome, checkpoint-SWAD with BatchNorm recalibration has higher mean
target accuracy and lower mean ECE, CwECE, and NLL than both Source-Acc and
AC-NC (Table~\ref{tab:swad-detailed}). These comparisons concern the complete
averaging-and-BatchNorm pipeline; they do not isolate a benefit from weight
averaging alone.

\paragraph{BatchNorm dependence.}
On OfficeHome, the no-recalibration variant has lower mean target accuracy
than the recalibrated variant (48.36 versus 62.62) and a much higher mean NLL
(9567.3991 versus 2.0026). Its NLL exceeds 100 in 16 of 180 runs, while its
median NLL is 1.9903. These severe cases arise from mismatches between the
averaged parameters and the buffers of the first selected checkpoint. Thus
the no-recalibration result should be interpreted as a buffer mismatch
diagnostic, not as an exact reproduction of SWAD with BatchNorm frozen
throughout training.

\FloatBarrier

\subsection{LODO validation details}
\label{app:lodo-details}

\paragraph{Protocol.}
The LODO study covers ERM and VREx on OfficeHome and TerraIncognita. For each
dataset, algorithm, hyperparameter seed, and trial seed, an inner run excludes
an unordered pair of domains and trains on the remaining two. It evaluates
both excluded domains at steps $0,100,\ldots,5000$. For an outer target domain
$T$, only the score of the other excluded domain is used as a pseudo-target
validation signal; the true target $T$ is never used for checkpoint or
hyperparameter selection. Reusing each unordered pair for both orderings gives
108 inner runs per algorithm and 216 in total.

Source-Acc and AC use deterministic re-evaluation of the full-source
trajectory. LODO-Acc instead maximizes mean pseudo-target accuracy across the
three inner folds associated with the outer target. AC-LODO first retains
checkpoints within 0.5 percentage points of the best LODO accuracy and then
ranks them using LODO NLL and CwECE with the same normalized $D_\infty$ rule as
AC. All four selectors use 15-bin hard ECE, absolute calibration gaps, and
class-frequency-weighted CwECE; NLL is computed directly from logits. These
calibration values are therefore not numerically comparable with the Gaussian
soft-bin squared-gap metrics of the main cohort.

We report two scopes. The fixed-hyperparameter scope evaluates all nine
hyperparameter-seed and trial-seed blocks, with four held-out domains per block,
giving 36 target runs per dataset, algorithm, and selector. Its intervals use
10,000 paired bootstrap resamples of the nine blocks. The hyperparameter-selection
scope selects among the three hyperparameter seeds within each trial, leaving
only three trial blocks and 12 target runs per dataset, algorithm, and selector;
its intervals are descriptive.

\begin{table}[ht]
\centering
\caption{Fixed-hyperparameter LODO results (36 runs per row). Accuracy is in percent; hard-bin ECE and class-frequency-weighted CwECE are multiplied by 100; NLL is unscaled.}
\label{tab:lodo-fixed-detailed}
\small
\setlength{\tabcolsep}{3.5pt}
\begin{tabular}{lllrrrr}
\toprule
Algorithm & Dataset & Selector & Acc. $\uparrow$ & NLL $\downarrow$ & ECE $\downarrow$ & CwECE $\downarrow$ \\
\midrule
ERM & OfficeHome & Source-Acc & 65.786 & 1.501 & 11.661 & 0.716 \\
 & & AC & 65.636 & 1.507 & 11.603 & 0.719 \\
 & & LODO-Acc & 65.754 & 1.447 & 9.349 & 0.699 \\
 & & AC-LODO & 65.720 & 1.443 & 8.621 & 0.703 \\
\cmidrule(lr){2-7}
 & TerraIncognita & Source-Acc & 46.199 & 2.487 & 32.301 & 13.131 \\
 & & AC & 45.809 & 2.446 & 32.070 & 13.002 \\
 & & LODO-Acc & 46.150 & 2.200 & 28.551 & 12.552 \\
 & & AC-LODO & 45.893 & 2.190 & 28.630 & 12.505 \\
\midrule
VREx & OfficeHome & Source-Acc & 57.557 & 1.935 & 13.719 & 0.739 \\
 & & AC & 57.585 & 1.933 & 13.810 & 0.739 \\
 & & LODO-Acc & 57.235 & 1.924 & 12.426 & 0.742 \\
 & & AC-LODO & 57.073 & 1.918 & 12.097 & 0.745 \\
\cmidrule(lr){2-7}
 & TerraIncognita & Source-Acc & 40.464 & 2.249 & 26.176 & 13.224 \\
 & & AC & 40.763 & 2.252 & 26.234 & 13.227 \\
 & & LODO-Acc & 39.870 & 2.079 & 22.378 & 12.904 \\
 & & AC-LODO & 40.071 & 2.022 & 21.106 & 12.874 \\
\bottomrule
\end{tabular}
\end{table}

\begin{table}[ht]
\centering
\caption{Paired LODO contrasts, averaged equally over the two algorithms and two datasets. Each entry is the mean difference followed by a 95\% bootstrap interval. Positive accuracy and negative error differences favor the first selector. Hyperparameter-selection intervals are descriptive because only three trial blocks are available.}
\label{tab:lodo-combined-contrasts}
\small
\setlength{\tabcolsep}{4pt}
\begin{tabular}{llcc}
\toprule
Scope & Metric & AC $-$ LODO-Acc & AC-LODO $-$ LODO-Acc \\
\midrule
Fixed & Accuracy & $+0.196\;[-0.755,+1.291]$ & $-0.062\;[-0.441,+0.367]$ \\
 & NLL & $+0.122\;[+0.061,+0.183]$ & $-0.020\;[-0.055,+0.005]$ \\
 & ECE & $+2.753\;[+1.837,+3.681]$ & $-0.562\;[-1.160,-0.099]$ \\
 & CwECE & $+0.198\;[-0.064,+0.468]$ & $-0.017\;[-0.130,+0.077]$ \\
\midrule
Selected & Accuracy & $-0.485\;[-1.206,+0.473]$ & $+0.175\;[-0.074,+0.486]$ \\
 & NLL & $+0.286\;[+0.188,+0.388]$ & $-0.032\;[-0.064,-0.004]$ \\
 & ECE & $+6.041\;[+4.332,+7.635]$ & $-0.789\;[-1.189,-0.395]$ \\
 & CwECE & $+0.823\;[+0.284,+1.355]$ & $-0.138\;[-0.321,+0.002]$ \\
\bottomrule
\end{tabular}
\end{table}

\paragraph{Effect of LODO validation.}
In the fixed-hyperparameter scope, AC and LODO-Acc have no resolved accuracy
difference after equal weighting over algorithms and datasets
($+0.196$ percentage points; 95\% CI: $[-0.755,+1.291]$). However, AC has
higher NLL by $0.122$ ($[+0.061,+0.183]$) and higher ECE by $2.753$ on the
$\times100$ scale ($[+1.837,+3.681]$). This pattern is directionally
consistent for ERM and VREx and indicates that the pseudo-target LODO signal
selects checkpoints with better target reliability in this experiment. The
comparison changes the validation protocol as well as the selected checkpoint,
so it does not isolate reliability ranking under a common feasible set.

\paragraph{Reliability ranking within LODO.}
Relative to LODO-Acc, AC-LODO changes fixed-hyperparameter accuracy by
$-0.062$ percentage points ($[-0.441,+0.367]$) and ECE by $-0.562$
($[-1.160,-0.099]$). The NLL and CwECE intervals include zero. The ECE
effect is stronger for VREx ($-0.800$, $[-1.892,-0.017]$) than for ERM
($-0.324$, $[-0.835,+0.132]$). Thus the fixed-hyperparameter results support
an ECE reduction from reliability ranking inside the LODO feasible set, but
not an accuracy improvement or uniform improvement across all reliability
metrics.

The hyperparameter-selection scope has the same qualitative pattern. AC-LODO
minus LODO-Acc changes accuracy by $+0.175$ percentage points
($[-0.074,+0.486]$), NLL by $-0.032$ ($[-0.064,-0.004]$), and ECE by
$-0.789$ ($[-1.189,-0.395]$). Because this scope has only three trial blocks
per algorithm and dataset, these intervals describe the evaluated trials and
are not used as stable population-level uncertainty statements. Across both
scopes, the 0.5-point constraint applies to source or LODO validation
accuracy, not to target accuracy. The evidence is limited to ERM and VREx on
OfficeHome and TerraIncognita.

\FloatBarrier

\subsection{VREx-only PAIR-s-style comparison}
\label{sec:pair-diagnostic}

Table~\ref{tab:pairs} compares the two PAIR-s-style selectors with source-out
accuracy and the default AC selector. This diagnostic is restricted to
VREx because the required penalty is unavailable for the other training
methods. It is not a reproduction of PAIR under its original training setup.

\begin{table}[ht]
\centering
\caption{VREx-only comparison over 36 runs per dataset. Acc., WHD, and WC are percentages; ECE and CwECE are multiplied by 100, while NLL retains its original scale. WHD and WC denote worst held-out-domain and worst-class accuracy, respectively.}
\label{tab:pairs}
\resizebox{\linewidth}{!}{%
\begin{tabular}{llrrrrrr}
\toprule
Dataset & Selector & Acc. (\%) $\uparrow$ & ECE ($\times100$) $\downarrow$ & CwECE ($\times100$) $\downarrow$ & NLL $\downarrow$ & WHD (\%) $\uparrow$ & WC (\%) $\uparrow$ \\
\midrule
OfficeHome & Source-Acc & 57.44 & 2.77 & 2.92 & 1.9327 & 45.61 & 12.08 \\
& PAIR iid-last10 & 50.03 & 2.11 & 3.38 & 2.2105 & 38.58 & 10.92 \\
& PAIR val-filter & 57.44 & 2.94 & 3.12 & 1.9435 & 45.95 & 11.93 \\
& AC-NC & 57.63 & 2.54 & 2.70 & 1.9025 & 45.49 & 12.00 \\
\midrule
PACS & Source-Acc & 79.79 & 2.05 & 3.17 & 0.7219 & 70.53 & 56.69 \\
& PAIR iid-last10 & 70.12 & 1.88 & 2.26 & 0.9669 & 62.25 & 49.57 \\
& PAIR val-filter & 80.00 & 1.95 & 3.14 & 0.7098 & 71.10 & 56.89 \\
& AC-NC & 80.32 & 1.91 & 2.99 & 0.6787 & 71.79 & 58.73 \\
\midrule
TerraInc. & Source-Acc & 41.74 & 8.49 & 10.36 & 2.2432 & 32.90 & 0.20 \\
& PAIR iid-last10 & 39.95 & 8.07 & 9.53 & 2.3009 & 31.65 & 0.34 \\
& PAIR val-filter & 41.39 & 8.73 & 10.26 & 2.2444 & 32.29 & 0.50 \\
& AC-NC & 41.77 & 8.02 & 10.11 & 2.1788 & 33.74 & 0.30 \\
\bottomrule
\end{tabular}
}
\end{table}
\FloatBarrier

\section{Ablations and Selection Diagnostics}
\label{app:ablations}

The following analyses isolate the choices internal to AC and characterize
when they matter. We proceed from the post-development distance comparison to
objective-set and component ablations, tolerance sensitivity, feasible-set
geometry, calibration-only failure cases, and local source--target ranking
diagnostics. Target-domain quantities in this section are evaluation
diagnostics and never enter checkpoint selection.

\subsection{Post-development evaluation and direct distance comparisons}
\label{sec:validation-results}

Table~\ref{tab:validation} restricts evaluation to OfficeHome and TerraIncognita, on which the NC objectives and $\delta=0.5$ were fixed after development on PACS. All three distances reduce the mean target calibration errors and NLL; their accuracy intervals include zero. The distance choice was not fixed before these evaluations, so distance-specific comparisons are exploratory. Table~\ref{tab:distance-direct} directly compares $D_\infty$ with $D_1$ at the same objectives and tolerance.

\begin{table}[ht]
\centering
\caption{Paired changes from Source-Acc on OfficeHome and TerraIncognita (360 runs). Accuracy differences are in percentage points; calibration differences are multiplied by 100. The second line gives 95\% percentile bootstrap intervals.}
\label{tab:validation}
\small\setlength{\tabcolsep}{4pt}
\begin{tabular}{lrrrr}
\toprule
NC distance & $\Delta$ Acc. & $\Delta$ ECE & $\Delta$ CwECE & $\Delta$ NLL\\
\midrule
$D_1$ & $+0.2209$ & $-0.2337$ & $-0.1611$ & $-0.0289$\\
 & $[-0.0464,+0.5072]$ & $[-0.3975,-0.0853]$ & $[-0.2640,-0.0579]$ & $[-0.0473,-0.0115]$\\
$D_2$ & $+0.2234$ & $-0.2428$ & $-0.1754$ & $-0.0296$\\
 & $[-0.0448,+0.5093]$ & $[-0.4062,-0.0930]$ & $[-0.2783,-0.0728]$ & $[-0.0480,-0.0120]$\\
$D_\infty$ & $+0.2133$ & $-0.2395$ & $-0.1821$ & $-0.0295$\\
 & $[-0.0592,+0.5017]$ & $[-0.4038,-0.0903]$ & $[-0.2841,-0.0803]$ & $[-0.0483,-0.0119]$\\
\bottomrule
\end{tabular}
\end{table}

\FloatBarrier

\begin{table}[ht]
\centering
\caption{NC $D_\infty$ minus NC $D_1$ on all 540 runs at $\delta=0.5$. Accuracy differences are in percentage points; calibration differences are multiplied by 100.}
\label{tab:distance-direct}
\begin{tabular}{lrr}
\toprule
Target metric & Mean difference & 95\% paired bootstrap CI\\
\midrule
Accuracy & $-0.0095$ & $[-0.0561,+0.0270]$\\
ECE & $-0.0065$ & $[-0.0270,+0.0122]$\\
CwECE & $-0.0200$ & $[-0.0462,+0.0024]$\\
NLL & $-0.0006$ & $[-0.0035,+0.0019]$\\
\bottomrule
\end{tabular}
\end{table}
\FloatBarrier

\subsection{Reliability-objective and distance ablations}
\label{app:aggregation-details}

Table~\ref{tab:six-variant} reports the aggregate comparison described in Section~\ref{sec:experiments} of the main paper. The objective sets have similar source-accuracy gaps, while their target reliability changes differ, particularly for NLL\@. The NC results are close across $D_1$, $D_2$, and $D_\infty$.
The archived NC/NE/NEC selector applied Pareto filtering after normalization. Removing it changed no selected checkpoint in the reported objective-set, distance, or tolerance ablations; the main paper therefore states the equivalent direct distance rule for these results.

\begin{table}[ht]
\centering
\caption{All nine objective-set and aggregation combinations at $\delta=0.5$, as paired mean changes from Source-Acc over 540 runs. Accuracy and WHD differences are in percentage points; calibration differences are multiplied by 100.}
\label{tab:six-variant}
\small
\begin{tabular}{llrrrrrr}
\toprule
Objectives & Distance & $\Delta$ src. acc. & $\Delta$ tgt. acc. & $\Delta$ ECE & $\Delta$ CwECE & $\Delta$ NLL & $\Delta$ WHD\\
\midrule
NC & $D_1$ & $-0.1455$ & $+0.0940$ & $-0.1645$ & $-0.0995$ & $-0.0213$ & $+0.1788$\\
NC & $D_2$ & $-0.1442$ & $+0.0848$ & $-0.1723$ & $-0.1109$ & $-0.0218$ & $+0.1745$\\
NC & $D_\infty$ & $-0.1450$ & $+0.0845$ & $-0.1710$ & $-0.1195$ & $-0.0219$ & $+0.1965$\\
NE & $D_1$ & $-0.1472$ & $+0.0159$ & $-0.1489$ & $-0.0844$ & $-0.1583$ & $+0.1483$\\
NE & $D_2$ & $-0.1469$ & $+0.0007$ & $-0.1438$ & $-0.0792$ & $-0.1574$ & $+0.1583$\\
NE & $D_\infty$ & $-0.1457$ & $-0.0001$ & $-0.1345$ & $-0.0762$ & $-0.1568$ & $+0.1677$\\
NEC & $D_1$ & $-0.1517$ & $+0.1232$ & $-0.2119$ & $-0.1530$ & $-0.1656$ & $+0.1870$\\
NEC & $D_2$ & $-0.1529$ & $+0.0877$ & $-0.1944$ & $-0.1602$ & $-0.1637$ & $+0.1974$\\
NEC & $D_\infty$ & $-0.1506$ & $+0.0889$ & $-0.1920$ & $-0.1382$ & $-0.1617$ & $+0.2778$\\
\bottomrule
\end{tabular}
\end{table}

At fixed $D_\infty$, Table~\ref{tab:objective-direct} compares NEC directly
with NC on matched runs. The 95\% percentile intervals use 10,000 paired
run-level bootstrap resamples (seed 20260924). All four intervals include zero.
The NLL mean is sensitive to one OfficeHome/IRM run, whose NEC-minus-NC NLL
difference is $-73.1787$; removing this run changes the pooled mean to
$-0.0043$. The two selectors choose the same checkpoint in 491/540 runs.

\begin{table}[ht]
\centering
\caption{NEC minus NC at fixed $D_\infty$ and $\delta=0.5$ on 540 paired runs. Accuracy is in percentage points, calibration differences are multiplied by 100, and NLL is unscaled.}
\label{tab:objective-direct}
\begin{tabular}{lrr}
\toprule
Target metric & Mean difference & 95\% paired bootstrap CI\\
\midrule
Accuracy & $+0.0044$ & $[-0.0827,+0.0974]$\\
ECE & $-0.0210$ & $[-0.0653,+0.0186]$\\
CwECE & $-0.0187$ & $[-0.0560,+0.0165]$\\
NLL & $-0.1398$ & $[-0.4150,+0.0002]$\\
\bottomrule
\end{tabular}
\end{table}
\FloatBarrier

\subsection{Single-objective component ablation}
\label{sec:component-ablation}

AC-NLL and AC-CwECE apply the main-paper AC rule with singleton objective sets $\{\mathrm{NLL}\}$ and $\{\mathrm{CwECE}\}$. Both share the source-accuracy feasible set, domain averaging, and tie-breaking rule of AC-NC. Table~\ref{tab:component-source} gives source objectives and secondary target outcomes. AC-NLL attains the lowest source NLL, and AC-CwECE attains the lowest source CwECE. In the pooled source metrics, AC-NC lies between AC-NLL and AC-CwECE. This source-side trade-off does not establish a target-domain advantage.

\begin{table}[ht]
\centering
\caption{Source selection behavior and secondary target outcomes for the component ablation. Each rule uses the same feasible sets, with 3.920370 candidates on average. Gap is in percentage points; accuracy, CwECE, WHD, and worst-class accuracy (WC) are multiplied by 100. WHD averages 135 configuration-level minima.}
\label{tab:component-source}
\small
\setlength{\tabcolsep}{4pt}
\begin{tabular}{lrrrrrr}
\toprule
Selector & Gap & Src. acc. $\uparrow$ & Src. NLL $\downarrow$ & Src. CwECE $\downarrow$ & WHD $\uparrow$ & WC $\uparrow$\\
\midrule
AC-NLL & 0.138017 & 84.4905 & 1.269184 & 2.2609 & 51.5809 & 24.5767\\
AC-CwECE & 0.155944 & 84.4725 & 1.292866 & 2.1405 & 51.5854 & 24.5787\\
AC-NC & 0.145010 & 84.4835 & 1.271047 & 2.1888 & 51.6157 & 24.6432\\
\bottomrule
\end{tabular}
\end{table}

\FloatBarrier

Table~\ref{tab:component-dataset} separates the dataset outcomes. On PACS, AC-NC has higher mean target accuracy and lower ECE, CwECE, and NLL than either single-objective rule. OfficeHome results are close to AC-CwECE. On TerraIncognita, both single-objective rules have higher mean target accuracy and lower NLL than AC-NC, and AC-CwECE also has lower calibration errors.

\begin{table}[ht]
\centering
\caption{Component ablation by dataset (180 runs each, $\delta=0.5$). Gap is in percentage points; target accuracy, ECE, CwECE, WHD, and WC are multiplied by 100. NLL retains its original scale.}
\label{tab:component-dataset}
\small
\setlength{\tabcolsep}{4pt}
\begin{tabular}{llrrrrrrr}
\toprule
Dataset & Selector & Gap & Acc. $\uparrow$ & ECE $\downarrow$ & CwECE $\downarrow$ & NLL $\downarrow$ & WHD $\uparrow$ & WC $\uparrow$\\
\midrule
OfficeHome & AC-NLL & 0.1432 & 61.0523 & 2.5276 & 3.0798 & 4.240746 & 48.2747 & 12.8679\\
 & AC-CwECE & 0.1686 & 61.0611 & 2.4858 & 3.0615 & 4.236254 & 48.2056 & 13.0297\\
 & AC-NC & 0.1612 & 61.0613 & 2.4828 & 3.0488 & 4.236244 & 48.2139 & 12.9993\\
\midrule
PACS (dev.) & AC-NLL & 0.1689 & 80.3712 & 1.6402 & 2.8355 & 0.778687 & 71.8626 & 60.1070\\
 & AC-CwECE & 0.1660 & 80.5422 & 1.6172 & 2.8025 & 0.780122 & 72.1708 & 59.8490\\
 & AC-NC & 0.1647 & 80.5523 & 1.5924 & 2.7501 & 0.772756 & 72.1615 & 60.0456\\
\midrule
TerraInc. & AC-NLL & 0.1019 & 43.0980 & 8.8828 & 11.2735 & 2.311322 & 34.6055 & 0.7551\\
 & AC-CwECE & 0.1332 & 43.1120 & 8.8480 & 11.1724 & 2.309663 & 34.3797 & 0.8574\\
 & AC-NC & 0.1091 & 42.9749 & 8.9034 & 11.2546 & 2.320975 & 34.4716 & 0.8846\\
\bottomrule
\end{tabular}
\end{table}

\FloatBarrier

Table~\ref{tab:component-ci} reports 10,000 paired bootstrap resamples. Source and target accuracy, calibration, NLL, and worst-class accuracy use 540 matched run pairs. WHD accuracy uses 135 dataset--algorithm--hyperparameter--trial blocks, each containing the minimum target accuracy across four held-out domains. These intervals describe checkpoint reselection on the fixed trajectories. All paired 95\% target-metric intervals include zero; these comparisons establish neither superiority nor equivalence of the joint rule.

\begin{table}[ht]
\centering
\caption{Paired component differences (AC-NC minus each comparator), with 95\% percentile bootstrap intervals. Accuracy differences are in percentage points; calibration differences are multiplied by 100; NLL retains its original scale.}
\label{tab:component-ci}
\small
\setlength{\tabcolsep}{3pt}
\begin{tabular}{lrrrr}
\toprule
& \multicolumn{2}{c}{AC-NC minus AC-NLL} & \multicolumn{2}{c}{AC-NC minus AC-CwECE}\\
\cmidrule(lr){2-3}\cmidrule(lr){4-5}
Metric & Mean $\Delta$ & 95\% CI & Mean $\Delta$ & 95\% CI\\
\midrule
Source accuracy & $-0.006993$ & $[-0.018585,+0.004787]$ & $+0.010934$ & $[+0.000081,+0.022039]$\\
Target accuracy & $+0.022297$ & $[-0.106108,+0.145756]$ & $-0.042315$ & $[-0.185333,+0.102861]$\\
ECE & $-0.0240$ & $[-0.0900,+0.0460]$ & $+0.0092$ & $[-0.0942,+0.1303]$\\
CwECE & $-0.0424$ & $[-0.0948,+0.0097]$ & $+0.0084$ & $[-0.0590,+0.0753]$\\
NLL & $-0.000260$ & $[-0.008409,+0.008276]$ & $+0.001312$ & $[-0.011792,+0.016416]$\\
Worst-class accuracy & $+0.066479$ & $[-0.226859,+0.376962]$ & $+0.064450$ & $[-0.194531,+0.312541]$\\
WHD accuracy & $+0.034720$ & $[-0.188658,+0.231663]$ & $+0.030314$ & $[-0.281091,+0.324917]$\\
\bottomrule
\end{tabular}
\end{table}

\FloatBarrier

\subsubsection{Selector-disagreement counts}
\label{app:component-details}

AC-ECE, AC-NLL, AC-CwECE, and AC-NC use the same
$\Theta_{0.5}$ and identical tie-breaking, so differences between these
selectors arise only from the reliability-ranking objective.

Among the 391 runs with multiple eligible checkpoints, AC-NC selects a
different checkpoint from AC-NLL in 110 runs, from AC-CwECE in 121 runs, and
from AC-ECE in 207 runs. These disagreement counts show that the objectives
can induce different rankings, but they are not measures of target-domain
improvement.

Because every selector in this comparison shares the same feasible set, this
ablation also does not isolate the value of accuracy filtering itself. The
paired target-metric differences in Table~\ref{tab:component-main} of the main paper therefore address a
different question: whether the joint NLL+CwECE ranking is empirically
distinguished from accuracy-constrained single-objective rankings. The current
results do not establish such an advantage.

\subsection{Accuracy-tolerance ablation}
\label{sec:tolerance-ablation}
\label{app:tolerance-details}

We vary only $\delta$ for NC with $D_\infty$ on the same 540 trajectories. Table~\ref{tab:delta-nc-appendix} shows that increasing the tolerance from $0.1$ to $1.0$ expands the mean candidate set from 1.30 to 9.71 checkpoints. Relative to $\delta=0.5$, $\delta=0.1$ increases target ECE by $0.1314$ on the $\times100$ scale and NLL by $0.017667$, whereas $\delta=1.0$ reduces them by $0.1385$ on the $\times100$ scale and $0.332431$. The target-accuracy differences are $-0.0649$ and $-0.0190$ percentage points, respectively, and both confidence intervals include zero. Yet 74 and 55 runs, respectively, lose at least one target-accuracy point; the $\delta=1.0$ NLL interval also includes zero. Varying $\delta$ changes both the feasible set and the within-set normalization.

\begin{table}[ht]
\centering
\caption{NC $D_\infty$ tolerance ablation on 540 runs. Gap is the mean source-accuracy deficit in percentage points. Accuracy and calibration values are multiplied by 100.}
\label{tab:delta-nc-appendix}
\begin{tabular}{rrrrrrr}
\toprule
$\delta$ (pp) & Candidates & Gap (pp) & Acc. $\uparrow$ & ECE $\downarrow$ & CwECE $\downarrow$ & NLL $\downarrow$\\
\midrule
0.1 & 1.30 & 0.0063 & 61.46 & 4.46 & 5.78 & 2.4610\\
0.5 & 3.92 & 0.1450 & 61.53 & 4.33 & 5.69 & 2.4433\\
1.0 & 9.71 & 0.3516 & 61.51 & 4.19 & 5.62 & 2.1109\\
\bottomrule
\end{tabular}
\end{table}
\FloatBarrier

Table~\ref{tab:delta-by-dataset} resolves Table~\ref{tab:delta-nc-appendix} by dataset. PACS accuracy is highest at $\delta=0.1$, while TerraIncognita accuracy is highest at $\delta=1.0$ among the tested values.

\begin{table}[ht]
\centering
\caption{NC $D_\infty$ tolerance ablation by dataset (180 runs per dataset). Gap is in percentage points; accuracy and calibration values are multiplied by 100.}
\label{tab:delta-by-dataset}
\small
\begin{tabular}{llrrrrrr}
\toprule
Dataset & $\delta$ & Candidates & Gap & Acc. & ECE & CwECE & NLL\\
\midrule
OfficeHome & 0.1 & 1.28 & 0.0075 & 60.90 & 2.69 & 3.26 & 4.2633\\
 & 0.5 & 3.80 & 0.1612 & 61.06 & 2.48 & 3.05 & 4.2362\\
 & 1 & 9.44 & 0.4164 & 60.99 & 2.46 & 2.98 & 3.2860\\
PACS (dev.) & 0.1 & 1.40 & 0.0072 & 80.77 & 1.59 & 2.75 & 0.7741\\
 & 0.5 & 5.20 & 0.1647 & 80.55 & 1.59 & 2.75 & 0.7728\\
 & 1 & 14.23 & 0.3686 & 80.30 & 1.59 & 2.82 & 0.7766\\
TerraInc. & 0.1 & 1.21 & 0.0043 & 42.72 & 9.09 & 11.32 & 2.3455\\
 & 0.5 & 2.76 & 0.1091 & 42.97 & 8.90 & 11.25 & 2.3210\\
 & 1 & 5.46 & 0.2699 & 43.24 & 8.51 & 11.06 & 2.2701\\
\bottomrule
\end{tabular}
\end{table}
\FloatBarrier

\begin{table}[ht]
\centering
\caption{NC $D_\infty$ differences relative to $\delta=0.5$ on 540 run pairs. Accuracy differences are in percentage points; calibration differences are multiplied by 100. Intervals are 95\% paired percentile bootstrap intervals from 10,000 resamples (seed 20260924).}
\label{tab:delta-ci}
\begin{tabular}{llrr}
\toprule
Tolerance & Metric & Mean difference & 95\% CI\\
\midrule
0.1 & Source accuracy & $+0.1387$ & $[+0.1245,+0.1530]$\\
& Target accuracy & $-0.0649$ & $[-0.2724,+0.1364]$\\
& ECE & $+0.1314$ & $[+0.0317,+0.2375]$\\
& CwECE & $+0.0879$ & $[+0.0175,+0.1588]$\\
& NLL & $+0.017667$ & $[+0.0047,+0.0311]$\\
\midrule
1.0 & Source accuracy & $-0.2066$ & $[-0.2328,-0.1798]$\\
& Target accuracy & $-0.0190$ & $[-0.2011,+0.1554]$\\
& ECE & $-0.1385$ & $[-0.2683,-0.0288]$\\
& CwECE & $-0.0664$ & $[-0.1358,+0.0006]$\\
& NLL & $-0.332431$ & $[-1.0089,+0.0275]$\\
\bottomrule
\end{tabular}
\end{table}
\FloatBarrier

Reliability effects are not uniformly monotone across datasets. In particular,
PACS CwECE and NLL increase at $\delta=1.0$. The tested values therefore do
not identify a uniformly preferable tolerance. Changing $\delta$ affects both
candidate eligibility and the extrema used for feasible-set normalization.

\subsection{Feasible-set composition}
\label{app:feasible-details}

At $\delta=0.5$, the accuracy-feasible set is a singleton in 149 of the 540
original-cohort runs. The remaining 391 runs contain multiple eligible
checkpoints, and AC-NC differs from Source-Acc in 264 of them (67.5\%).
For NC, 347/540 runs have only one nondominated candidate, leaving 193 runs
with multiple candidates that exhibit a reliability trade-off. Among the 85
two-checkpoint feasible sets, 32 have opposing NLL and CwECE rankings; exact
within-set min--max scaling maps these pairs to $(0,1)$ and $(1,0)$, so all
three distances tie and select the same checkpoint under the common
tie-breaking rule.

\clearpage
\subsection{Pure calibration-only selection}

\begin{table}[ht]
\centering
\caption{Pure calibration-only selection can select degenerate low-accuracy checkpoints. Source and target accuracy are percentages; ECE and CwECE are multiplied by 100, while NLL retains its original scale. Step is the mean selected training step; parenthetical values report the percentage of runs selecting step 0.}
\label{tab:pure-collapse}
\small\setlength{\tabcolsep}{4pt}
\begin{tabular}{llrrrrrr}
\toprule
Dataset & Selector & Src. acc. & Tgt. acc. & ECE & CwECE & NLL $\downarrow$ & Step \\
\midrule
OfficeHome & Source-Acc & 76.14 & 60.87 & 2.77 & 3.32 & 4.2724 & 2553 \\
& pure ECE   & 2.96 & 2.55 & 0.04 & 0.04 & 4.1890 & 164 (94.4\%) \\
& pure CwECE & 2.07 & 1.91 & 0.03 & 0.01 & 4.2168 & 173 (95.6\%) \\
& pure NLL   & 74.54 & 60.40 & 1.92 & 2.82 & 1.7333 & 1723 \\
\midrule
PACS & Source-Acc & 93.68 & 80.73 & 1.63 & 2.75 & 0.7794 & 2656 \\
& pure ECE   & 85.10 & 73.58 & 0.94 & 2.54 & 0.7927 & 1264 (7.8\%) \\
& pure CwECE & 60.96 & 52.22 & 1.00 & 1.77 & 1.2627 & 1450 (41.1\%) \\
& pure NLL   & 92.73 & 80.02 & 1.49 & 2.77 & 0.6910 & 2176 \\
\midrule
TerraInc. & Source-Acc & 84.07 & 42.74 & 9.10 & 11.35 & 2.3438 & 3984 \\
& pure ECE   & 75.60 & 41.41 & 6.86 & 8.98 & 2.0836 & 2201 (6.1\%) \\
& pure CwECE & 53.27 & 28.24 & 3.74 & 4.08 & 2.0329 & 380 (29.4\%) \\
& pure NLL   & 83.27 & 42.78 & 8.90 & 11.28 & 2.2574 & 3909 \\
\bottomrule
\end{tabular}
\end{table}
\FloatBarrier

\subsection{Accuracy-plateau variation}
\label{app:plateau-variation}

For each run, Table~\ref{tab:plateau-variation} first measures the number of eligible checkpoints and the range of each metric inside $\Theta_{0.5}$, then averages these quantities within each dataset--algorithm group. 

\begin{table}[H]
\centering
\caption{Reliability variation within $\Theta_{0.5}$ (36 runs per row). Source-accuracy ranges are in percentage points; CwECE ranges use the $[0,1]$ scale; NLL ranges remain on their original scale.}
\label{tab:plateau-variation}
\small\setlength{\tabcolsep}{4pt}
\begin{tabular}{llrrrrrr}
\toprule
Dataset & Algorithm & Ckpts & \shortstack{Src. acc.\\range} & \shortstack{Src. NLL\\range} & \shortstack{Src. CwECE\\range} & \shortstack{Tgt. NLL\\range} & \shortstack{Tgt. CwECE\\range} \\
\midrule
OfficeHome & CORAL    & 4.33 & 0.3639 & 0.0689 & 0.0063 & 0.1498 & 0.0091 \\
           & ERM      & 4.03 & 0.3283 & 0.0441 & 0.0051 & 0.1196 & 0.0086 \\
           & GroupDRO & 5.08 & 0.3753 & 0.0642 & 0.0064 & 0.1691 & 0.0101 \\
           & IRM      & 2.11 & 0.0979 & 2.1954 & 0.0012 & 11.3938 & 0.0024 \\
           & VREx     & 3.44 & 0.2772 & 0.0388 & 0.0039 & 0.0852 & 0.0062 \\
\midrule
PACS       & CORAL    & 5.06 & 0.3727 & 0.0336 & 0.0032 & 0.2091 & 0.0138 \\
           & ERM      & 7.42 & 0.4347 & 0.0414 & 0.0027 & 0.2728 & 0.0177 \\
           & GroupDRO & 4.92 & 0.3611 & 0.0325 & 0.0031 & 0.2037 & 0.0152 \\
           & IRM      & 3.00 & 0.2203 & 0.2993 & 0.0023 & 0.2502 & 0.0073 \\
           & VREx     & 5.61 & 0.3416 & 0.0324 & 0.0033 & 0.1654 & 0.0112 \\
\midrule
TerraInc.  & CORAL    & 2.47 & 0.2270 & 0.0154 & 0.0045 & 0.2309 & 0.0155 \\
           & ERM      & 4.78 & 0.3271 & 0.0224 & 0.0121 & 0.4823 & 0.0330 \\
           & GroupDRO & 1.83 & 0.2021 & 0.0123 & 0.0029 & 0.1572 & 0.0128 \\
           & IRM      & 1.89 & 0.1332 & 1.4568 & 0.0045 & 0.1938 & 0.0073 \\
           & VREx     & 2.83 & 0.2338 & 0.0166 & 0.0059 & 0.3176 & 0.0185 \\
\bottomrule
\end{tabular}
\end{table}
\FloatBarrier

\subsection{Source--target rank correlation}

Table~\ref{tab:rank-corr} compares rank correlation within $\Theta_{0.5}$.
Each NC utility is the negative $D_\infty$ of NLL and CwECE normalized over
that same feasible set, using source or target measurements, respectively.
Only runs with at least three feasible checkpoints and nonconstant utilities
are evaluable; Spearman correlation uses average ranks for ties. The source
accuracy proxy is instead compared with target accuracy. These correlations
describe local ranking, not target-optimal selection.

\begin{table}[ht]
\centering
\caption{Mean within-$\Theta_{0.5}$ rank correlation. Source accuracy is paired with target accuracy; source NC utility is paired with target NC utility. Valid runs have at least three feasible checkpoints and nonconstant utilities; ties receive average ranks for Spearman $\rho$.}
\label{tab:rank-corr}
\resizebox{\linewidth}{!}{%
\begin{tabular}{llrrrr}
\toprule
Dataset & Source proxy & Spearman $\rho$ $\uparrow$ & Kendall $\tau$ $\uparrow$ & Checkpoints & Valid/total \\
\midrule
OfficeHome & source-out accuracy & $-0.045$ & $-0.059$ & 684 & 104/180 \\
& AC utility & $+0.513$ & $+0.438$ & 684 & 104/180 \\
PACS & source-out accuracy & $+0.138$ & $+0.122$ & 936 & 128/180 \\
& AC utility & $+0.092$ & $+0.074$ & 936 & 128/180 \\
TerraInc. & source-out accuracy & $+0.065$ & $+0.068$ & 497 & 74/180 \\
& AC utility & $+0.114$ & $+0.113$ & 497 & 74/180 \\
\bottomrule
\end{tabular}
}
\end{table}
\FloatBarrier

\section{Theoretical Guarantees and Proofs}
\label{app:theory}
\label{sec:selection-theory}

The theoretical results separate three claims that should not be conflated:
source-only statistics cannot guarantee target-optimal selection for
unrestricted targets; source--target ordering can transfer under an explicit
conditional structure; and the AC tolerance controls population source
accuracy for a fixed, validation-independent checkpoint set. The final
subsection extends the source-accuracy bound to arbitrary fixed domain
weights.

\subsection{Proof of Proposition 1}
\label{app:prop1-proof}

Recall that the source observations and saved trajectory are fixed.
Suppose the trajectory contains checkpoints $\theta_a$ and $\theta_b$ such
that
\[
f_{\theta_a}(x)\ne f_{\theta_b}(x)
\]
for some input $x$.

Consider two possible target distributions. The first is a point mass on
\[
\bigl(x,f_{\theta_a}(x)\bigr),
\]
and the second is a point mass on
\[
\bigl(x,f_{\theta_b}(x)\bigr).
\]
Under the first target distribution, any checkpoint predicting
$f_{\theta_a}(x)$ is target-accuracy optimal, whereas under the second, any
checkpoint predicting $f_{\theta_b}(x)$ is optimal. Since the two labels are
different, these two sets of target-accuracy maximizers are disjoint.

The source observations and saved trajectory are identical under the two
possible target distributions. Consequently, any source-only selector has the
same output distribution in both cases. That output distribution cannot assign
probability one to two disjoint sets of checkpoints. Hence no source-only
selector can select a target-accuracy maximizer with probability one for every
unrestricted target distribution.

This argument applies to deterministic and randomized selectors, including
Source-Acc and AC. It is specifically an impossibility statement for target
accuracy under unrestricted targets; it does not imply that source statistics
are uninformative under additional assumptions, nor does it establish an
analogous impossibility result for every reliability metric.

\subsection{Conditional source--target ordering transfer}
\label{app:conditional-transfer}

The impossibility result above does not exclude source--target ranking transfer
under additional structure. We record a conditional population-level
comparison here. This result is descriptive rather than operational: the
target-dependent quantities introduced below are unobserved and are not used
by AC.

For metric $m$ and
$d\in\{\mathrm{src},\mathrm{tgt}\}$, let
$U_{d,m}(\theta)$ denote population utility, with larger values preferred.
Accuracy is used directly, whereas NLL, ECE, and CwECE may equivalently be
represented by their negatives.

\paragraph{Monotone transfer decomposition.}
Fix a strictly increasing function $\psi_m$ on the source-utility values
attained by checkpoints in $\Theta$, and write
\begin{equation}
U_{\mathrm{tgt},m}(\theta)
=
\psi_m\!\left(U_{\mathrm{src},m}(\theta)\right)
+
\varepsilon_m(\theta),
\qquad
\theta\in\Theta.
\label{eq:app-transfer-decomposition}
\end{equation}
\label{eq:conditional-transfer-model}
No distributional assumption or magnitude bound is imposed on
$\varepsilon_m$. In particular, $\psi_m(u)=u$ is admissible, so
Eq.~\eqref{eq:app-transfer-decomposition} alone places no restriction on target
rankings.

For the fixed map $\psi_m$, define
\[
\omega_m(\Theta)
=
\max_{\theta_a,\theta_b\in\Theta}
\left|
\varepsilon_m(\theta_a)-\varepsilon_m(\theta_b)
\right|.
\]

\noindent\textbf{Proposition (conditional ordering transfer).}
For $\theta_i,\theta_j\in\Theta$, let
\[
u_i=U_{\mathrm{src},m}(\theta_i),
\qquad
u_j=U_{\mathrm{src},m}(\theta_j).
\]
If $u_i>u_j$, then
\begin{equation}
\begin{aligned}
U_{\mathrm{tgt},m}(\theta_i)
-
U_{\mathrm{tgt},m}(\theta_j)
&=
\psi_m(u_i)-\psi_m(u_j)
+
\varepsilon_m(\theta_i)-\varepsilon_m(\theta_j)\\
&\ge
\psi_m(u_i)-\psi_m(u_j)-\omega_m(\Theta).
\end{aligned}
\label{eq:app-transfer-bound}
\end{equation}
\label{eq:conditional-ordering-bound}
Consequently, the source ordering transfers strictly whenever
\begin{equation}
\psi_m(u_i)-\psi_m(u_j)
>
\omega_m(\Theta).
\label{eq:app-transfer-condition}
\end{equation}

\noindent\textit{Proof.}
Equation~\eqref{eq:app-transfer-decomposition} gives
\[
U_{\mathrm{tgt},m}(\theta_i)
-
U_{\mathrm{tgt},m}(\theta_j)
=
\psi_m(u_i)-\psi_m(u_j)
+
\varepsilon_m(\theta_i)-\varepsilon_m(\theta_j).
\]
By definition of $\omega_m(\Theta)$,
\[
\varepsilon_m(\theta_i)-\varepsilon_m(\theta_j)
\ge
-\omega_m(\Theta),
\]
which yields Eq.~\eqref{eq:app-transfer-bound}. If
Eq.~\eqref{eq:app-transfer-condition} holds, the right-hand side is positive,
and therefore
\[
U_{\mathrm{tgt},m}(\theta_i)
>
U_{\mathrm{tgt},m}(\theta_j).
\]
\hfill$\square$

The condition is sufficient but not necessary. If $u_i=u_j$, the target-side
difference is entirely determined by the residuals. Moreover, this comparison
uses population source utilities and unobserved target-dependent quantities:
it does not certify target ordering from an empirical source margin.
Per-metric ordering transfer also does not imply preservation of a normalized
multi-objective ranking such as the one used by AC.

\subsection{Proof of the finite-sample source-accuracy bound}
\label{app:source-bound-proof}

We prove Proposition 2. Condition on the fixed candidate set
$\Theta$, with $T=|\Theta|$, generated independently of the source validation
sets.

For a fixed checkpoint $\theta$, write
\[
Z_{e,i}(\theta)
=
\mathbf 1\{f_\theta(X_{e,i})=Y_{e,i}\},
\]
where
$(X_{e,i},Y_{e,i})$, $i=1,\ldots,n_e$, are the validation samples from source
domain $e$. Then
\[
\widehat A_{\mathrm{src}}(\theta)
=
\frac{100}{E}
\sum_{e=1}^{E}
\frac{1}{n_e}
\sum_{i=1}^{n_e}Z_{e,i}(\theta),
\]
and
\[
A_{\mathrm{src}}(\theta)
=
\mathbb E[
\widehat A_{\mathrm{src}}(\theta)
].
\]

For a fixed $\theta$, the weighted summands are independent and each has range
length
\[
\frac{100}{E n_e}
\]
for a sample from domain $e$. Hoeffding's inequality therefore gives, for
$r>0$,
\[
\Pr\!\left(
\left|
\widehat A_{\mathrm{src}}(\theta)
-
A_{\mathrm{src}}(\theta)
\right|
>r
\right)
\le
2\exp\left(
-\frac{2r^2}
{
\frac{100^2}{E^2}
\sum_{e=1}^{E}\frac{1}{n_e}
}
\right).
\]
Applying a union bound over the $T$ fixed checkpoints gives
\[
\Pr\!\left(
\sup_{\theta\in\Theta}
\left|
\widehat A_{\mathrm{src}}(\theta)
-
A_{\mathrm{src}}(\theta)
\right|
>r
\right)
\le
2T\exp\left(
-\frac{2r^2}
{
\frac{100^2}{E^2}
\sum_{e=1}^{E}\frac{1}{n_e}
}
\right).
\]
Setting the right-hand side equal to $\alpha$ yields
\[
r_\alpha
=
100
\sqrt{
\frac{\log(2T/\alpha)}{2E^2}
\sum_{e=1}^{E}\frac{1}{n_e}
}.
\]
Thus, with probability at least $1-\alpha$,
\begin{equation}
\sup_{\theta\in\Theta}
\left|
\widehat A_{\mathrm{src}}(\theta)
-
A_{\mathrm{src}}(\theta)
\right|
\le
r_\alpha.
\label{eq:app-uniform-accuracy-event}
\end{equation}

Let
\[
\theta^*
\in
\operatorname*{arg\,max}_{\theta\in\Theta}
A_{\mathrm{src}}(\theta)
\]
be a population source-accuracy maximizer, and let
$\theta\in\Theta_\delta$. On the event in
Eq.~\eqref{eq:app-uniform-accuracy-event},
\[
\begin{aligned}
A_{\mathrm{src}}(\theta)
&\ge
\widehat A_{\mathrm{src}}(\theta)-r_\alpha\\
&\ge
\widehat A_{\mathrm{src}}(\theta_{\mathrm{SA}})
-\delta-r_\alpha\\
&\ge
\widehat A_{\mathrm{src}}(\theta^*)
-\delta-r_\alpha\\
&\ge
A_{\mathrm{src}}(\theta^*)
-\delta-2r_\alpha.
\end{aligned}
\]
Since
$A_{\mathrm{src}}(\theta^*)
=
\max_{\theta'\in\Theta}A_{\mathrm{src}}(\theta')$,
this proves
\[
A_{\mathrm{src}}(\theta)
\ge
\max_{\theta'\in\Theta}A_{\mathrm{src}}(\theta')
-\delta-2r_\alpha.
\]
\hfill$\square$

The event in Eq.~\eqref{eq:app-uniform-accuracy-event} holds simultaneously
over the full fixed candidate set. Hence reliability-based ranking may reuse
the same source-validation samples after the feasible set is formed.
Independence between different checkpoints is not required. The essential
condition is that generation of the candidate trajectory be independent of
the validation sets to which the concentration argument is applied.
Validation feedback that changes candidate generation is not covered by this
guarantee.

\subsection{Weighted-domain extension of the source-accuracy bound}
\label{app:weighted-source-bound}

\noindent\textbf{Proposition (weighted finite-sample source-accuracy bound).}
Condition on $T$ checkpoints generated independently of the validation data.
For each of $E$ source domains, let $S_e$ contain $n_e$ i.i.d.\ observations
from $P_e$, with samples independent across domains. For fixed
$w_e\ge0$ satisfying $\sum_e w_e=1$, define
\[
A_w(\theta)=100\sum_{e=1}^{E}w_e\Pr_{P_e}(f_\theta(X)=Y),
\]
\[
\widehat A_w(\theta)=100\sum_{e=1}^{E}\frac{w_e}{n_e}
\sum_{(x,y)\in S_e}\mathbf 1\{f_\theta(x)=y\}.
\]
Let $\widehat\Theta_\delta$ contain checkpoints whose $\widehat A_w$
is within $\delta\ge0$ percentage points of its maximum.
For $\alpha\in(0,1)$, with probability at least $1-\alpha$, every
$\theta\in\widehat\Theta_\delta$ satisfies
\begin{equation}
A_w(\theta)\ge\max_{\theta'\in\Theta}A_w(\theta')-\delta
-200\sqrt{\frac{\log(2T/\alpha)}{2}
\sum_{e=1}^{E}\frac{w_e^2}{n_e}}.
\label{eq:weighted-source-regret}
\end{equation}

\noindent\textit{Proof.}
Weighted Hoeffding concentration and a union bound give
$|\widehat A_w(\theta)-A_w(\theta)|\le r$ simultaneously for all checkpoints,
where $r=100\sqrt{\frac{1}{2}\log(2T/\alpha)\sum_e w_e^2/n_e}$.
For each feasible checkpoint,
\[
A_w(\theta)\ge\widehat A_w(\theta)-r
\ge\max_{\theta'}\widehat A_w(\theta')-\delta-r
\ge\max_{\theta'}A_w(\theta')-\delta-2r.
\]
\noindent\hfill$\square$

The main-paper bound uses $w_e=1/E$. Weighting by sample count,
$w_e=n_e/\sum_j n_j$, recovers the bound with total sample size $\sum_e n_e$.
The result requires the checkpoint set to be generated independently of the
validation samples, but allows arbitrary dependence among checkpoints. Its
simultaneous guarantee also covers selection within $\widehat\Theta_\delta$
using reliability measured on the same samples. It concerns population source
accuracy, not target risk.

\section{Reproducibility and Figure Details}
\label{app:exp-details}

This final section records the inferential units, figure-generation
definitions, and retained training-code settings needed to interpret or
reproduce the reported analyses.

\subsection{Uncertainty and resampling conventions}
\label{app:exp-inference}

For the original checkpoint-selection cohort, uncertainty intervals are
computed from paired run-level differences. Main 95\% percentile intervals use
10,000 bootstrap resamples of the 360 post-development run pairs or the pooled
540 run pairs, depending on the analysis.

These intervals condition on the observed training trajectories. Runs that
share hyperparameter or trial seeds are not treated through an explicit
hierarchical dependence model in the main bootstrap. Accordingly, an interval
for the target-accuracy difference that contains zero should not be interpreted
as evidence that the two selectors are accuracy-equivalent.

The independent OfficeHome checkpoint-SWAD comparison is performed only within
its design-matched 180-run cohort; its rows are not paired with the
original-cohort rows in Table~\ref{tab:main}.

Worst-held-out-domain summaries and the remaining uncertainty analyses use
the aggregation and resampling conventions given in
Section~\ref{sec:aggregation-details}.

\subsection{Aggregation and resampling details}
\label{sec:aggregation-details}

\paragraph{Worst held-out-domain accuracy.}
For each dataset, algorithm, and hyperparameter--trial configuration, the four leave-one-domain-out runs supply four target accuracies. We take their minimum, then average over the nine configurations and the dataset--algorithm groups. This gives 135 configuration-level minima for the full collection and 90 for OfficeHome and TerraIncognita. The original paired analysis resampled 15 dataset--algorithm averages, whereas the NC tolerance analysis resampled 135 configuration minima. These are distinct resampling units for the same point estimate.

\paragraph{Paired resampling.}
Each resampled unit contains all compared selectors. The distance and two-benchmark analyses use 10,000 resamples and random seed 20260908. The reference NC, NE, and NEC intervals retain their original 10,000-resample estimates; the tolerance intervals use 10,000 resamples and seed 20260924. All comparisons use the same saved trajectories.

\subsection{Figure definitions}
\label{sec:figure-details}

Figure~\ref{fig:selection-overview}(a) of the main paper shows OfficeHome with domain A held out, VREx, hyperparameter seed 0, and trial seed 0. Source-Acc selects step 3500; the NC, NE, and NEC configurations select steps 3400, 1600, and 1500, respectively. Their $D_2$ and $D_\infty$ choices coincide in this run. Panel (b) compares all selectors using the same three target reliability metrics. For each run $r$ and $m\in\{\mathrm{NLL},\mathrm{ECE},\mathrm{CwECE}\}$, min--max normalization is performed over all 51 saved checkpoints $\Theta_r$:
\[
z^{\mathrm{tgt}}_{r,m}(\theta)=\frac{m_{\mathrm{tgt}}(\theta)-\min_{\theta'\in\Theta_r}m_{\mathrm{tgt}}(\theta')}{\max_{\theta'\in\Theta_r}m_{\mathrm{tgt}}(\theta')-\min_{\theta'\in\Theta_r}m_{\mathrm{tgt}}(\theta')+10^{-12}}.
\]
The ordinate is the mean Euclidean distance of the selected checkpoints,
\[
\overline D_{2,\mathrm{tgt}}(s)=\frac{1}{540}\sum_{r=1}^{540}\left(\sum_m [z^{\mathrm{tgt}}_{r,m}(\widehat\theta_{r,s})]^2\right)^{1/2},
\]
where $s$ indexes the source-only selector. The abscissa is mean target accuracy in percent. NC, NE, and NEC use source-side $D_\infty$ and $\delta=0.5$; pure selectors minimize their source metric over the full trajectory, with ties resolved by the earliest step. These target distances are computed only after selection.

Figure~\ref{fig:ac-nc-across-methods} of the main paper summarizes
algorithm-level results and checkpoint-rank agreement. Panel (a) reports mean
paired changes in target accuracy and NLL relative to Source-Acc over the
pooled 540 trajectories. Open markers denote AC-NC/$D_2$, and filled markers
denote AC-NEC/$D_\infty$; 
Accuracy changes are reported in percentage points and NLL changes on the original scale.
Panel (b) reports source--target Kendall $\tau_b$ for checkpoint rankings of
accuracy, NLL, ECE, and CwECE by dataset and algorithm. The
within-feasible-set NC-utility correlations in Table~\ref{tab:rank-corr} use
a different ranking scope and utility definition.

\FloatBarrier

\subsection{Training-code settings}
\label{sec:training-settings}

\begin{table}[H]
\centering
\caption{Training-code defaults and logged trajectory steps for the main sweep.}
\label{tab:training-settings}
\footnotesize
\setlength{\tabcolsep}{4pt}
\begin{tabular}{>{\raggedright\arraybackslash}p{0.22\linewidth}>{\raggedright\arraybackslash}p{0.70\linewidth}}
\toprule
Item & Setting \\
\midrule
Backbone & Code default: ImageNet-1K-pretrained ResNet-50 (\texttt{resnet50.ram\_in1k}). \\
Source split & Code default: 20\% validation per source domain; target domain excluded from selection. \\
Trajectory & 5001 updates; logged steps $0,100,\ldots,5000$ (51 per run). \\
Seeds & Hyperparameter seed 0: defaults; 1--2: registry draws. Trial seeds 0--2: splits and, jointly with hyperparameter seeds, draws. Training seed hashes all run keys. \\
Shared search & Learning rate $10^{-5}$--$10^{-3.5}$; weight decay $10^{-6}$--$10^{-2}$; batch size 8--45; dropout $\{0,0.1,0.5\}$. \\
Algorithm search & CORAL penalty weight $\gamma\in[10^{-1},10^1]$; GroupDRO $\eta\in[10^{-3},10^{-1}]$; IRM/VREx penalty weight $10^{-1}$--$10^5$ and annealing step 1--9999. \\
Aggregation & Select within each run; average 540 runs equally; no cross-configuration selection. \\
\bottomrule
\end{tabular}
\end{table}

\FloatBarrier

\end{document}